\documentclass[runningheads]{llncs}

\usepackage{eccv}

\usepackage{eccvabbrv}

\usepackage{graphicx}
\usepackage{booktabs}
\usepackage{multirow}
\usepackage{float}
\usepackage{algorithm}
\usepackage{algpseudocode}
\newcommand{\noopsort}[1]{}

\usepackage[accsupp]{axessibility}  

\usepackage[pagebackref,breaklinks,colorlinks,citecolor=eccvblue]{hyperref}

\usepackage{orcidlink}

\begin{document}

\title{AVSR-Diff: Scale-Agnostic Diffusion Priors\\for Temporally Consistent Arbitrary-Scale\\Video Super-Resolution} 

\titlerunning{AVSR-Diff}

\author{Geunhyuk Youk\inst{1}\orcidlink{0009-0000-2674-7741} \and
Jeonghyeok Do\inst{1}\orcidlink{0000-0003-0030-0129} \and
Dayeon Kim\inst{1}\orcidlink{0000-0002-9758-4676} \\
Jihyong Oh\inst{2}\textsuperscript{\textdagger}\orcidlink{0000-0002-1627-0529} \and
Munchurl Kim\inst{1}\textsuperscript{\textdagger}\orcidlink{0000-0003-0146-5419}}

\authorrunning{G.~Youk et al.}

\institute{KAIST, Republic of Korea\\
\email{\{rmsgurkjg, ehwjdgur0913, dyk4501, mkimee\}@kaist.ac.kr} \and
CMLab, Chung-Ang University, Republic of Korea\\
\email{jihyongoh@cau.ac.kr}}

\maketitle

\renewcommand{\thefootnote}{}%
\footnotetext{\textsuperscript{\textdagger}~Co-corresponding authors.}
\renewcommand{\thefootnote}{\arabic{footnote}}%

\begin{center}
\vspace{-4mm}
\url{https://kaist-viclab.github.io/AVSR-Diff/}
\vspace{-3mm}
\end{center}

\begin{figure}[h!]
    \centering
    \setlength{\tabcolsep}{0.1cm}
    \begin{tabular}{cc}
        \includegraphics[width=0.59\linewidth]{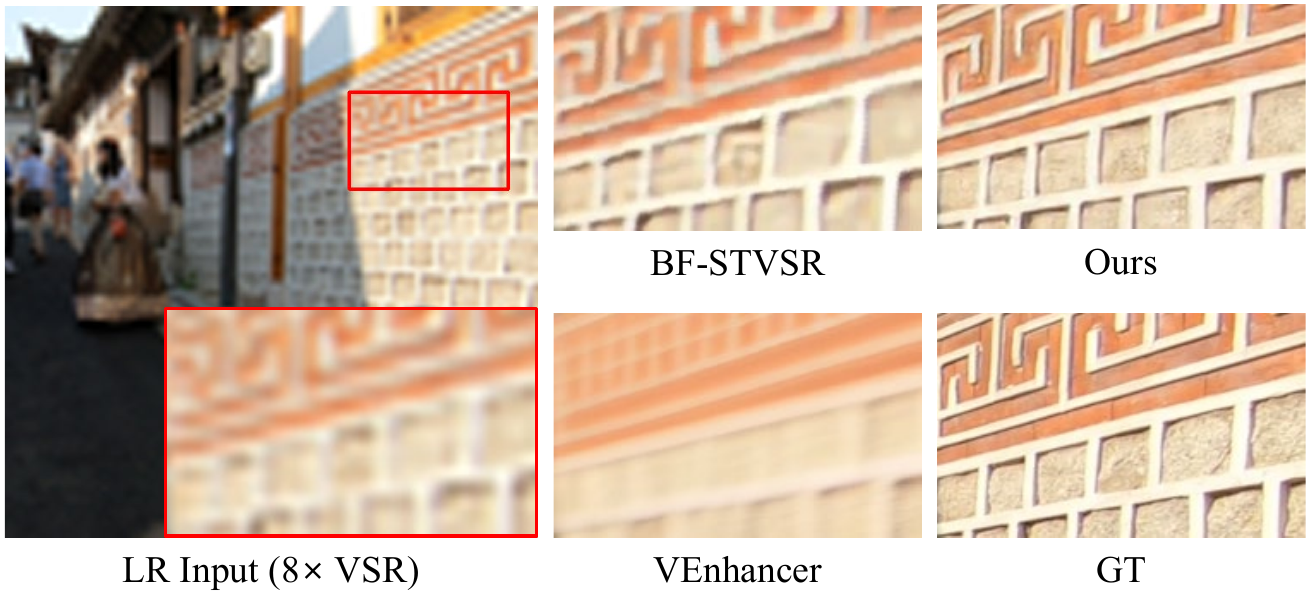} &
        \includegraphics[width=0.39\linewidth]{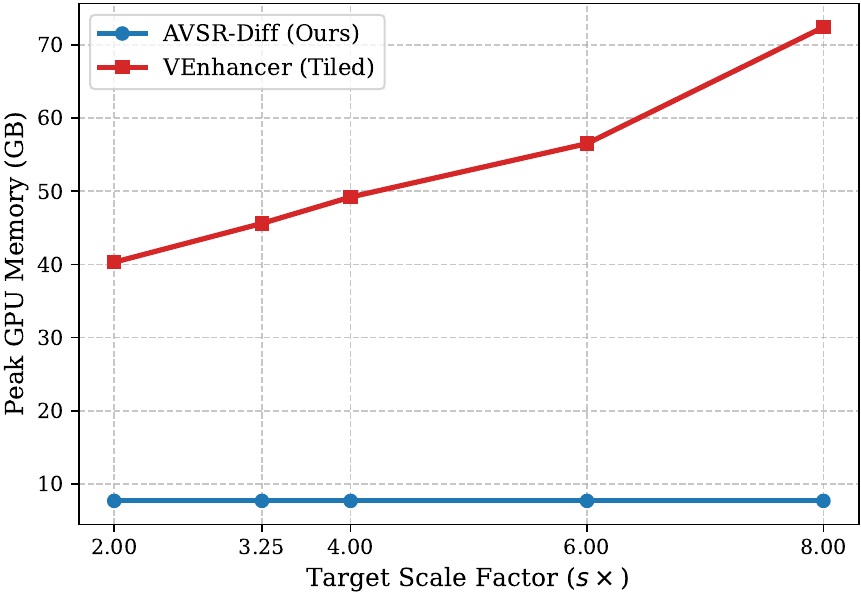} \\
        \multicolumn{1}{p{0.58\linewidth}}{\centering \footnotesize (a) Qualitative comparison at $8\times$ scale.} &
        \multicolumn{1}{p{0.4\linewidth}}{\centering \footnotesize (b) Peak GPU memory on a 10-frame $180\times320$ sequence.}
    \end{tabular}
    \vspace{-2mm}
    \caption{AVSR-Diff outperforms state-of-the-art methods in visual quality at large scale while maintaining a highly efficient, constant memory footprint.}
    \vspace{-12mm}
    \label{fig:teaser}
\end{figure}

\begin{sloppypar}
\begin{abstract}
Diffusion models have significantly advanced video super-resolution (VSR) but remain largely constrained to fixed upsampling scales. Conversely, while coordinate-based arbitrary-scale VSR methods offer scale flexibility, they inherently suffer from severe over-smoothing at large scaling factors. Integrating generative priors with continuous decoding is promising but currently hindered by severe temporal flickering caused by the stochasticity of diffusion sampling. To address this, we propose AVSR-Diff (Arbitrary-scale Video Super-Resolution with Diffusion), a novel decoupled framework that separates scale-agnostic latent denoising from continuous coordinate rendering, effectively avoiding computationally heavy resolution-specific sampling. Our approach introduces a Temporally-Gated Feature Recurrence (TGFR) module to extract strictly aligned, temporally consistent latent priors. Furthermore, we design a continuous video VAE decoder incorporating a Scale-Aware Fourier Refinement (SAFR) module to dynamically adapt frequency components to any target scale. Extensive experiments demonstrate that AVSR-Diff consistently preserves high-frequency details and strong temporal stability across various scales, surpassing state-of-the-art arbitrary-scale baselines. Remarkably, our framework outperforms recent fixed-scale generative models even on their native resolution.
\end{abstract}

\section{Introduction}
\label{sec:intro}
Video super-resolution (VSR) aims to restore high-resolution (HR) videos from low-resolution (LR) inputs by leveraging temporal information across adjacent frames. While conventional deep learning-based VSR methods~\cite{chan2021basicvsr, chan2022basicvsr++, liang2022rvrt, liang2024vrt, xu2024iart, liu2022ttvsr, tian2020tdan} have achieved impressive reconstruction accuracy, they often suffer from over-smoothed textures and fail to synthesize realistic high-frequency details. To overcome this perception-distortion trade-off, recent diffusion model (DM)-based approaches~\cite{rota2024stablevsr, zhou2024upscale, yang2024mgldvsr, xie2025star, xu2025dgafvsr, chen2025dove} have brought powerful generative priors to the VSR task, significantly improving perceptual quality. Despite their impressive generative capabilities, most existing DM-based VSR models are designed and trained for a \textit{fixed integer scaling factor} (typically $4\times$). Given that practical scenarios frequently demand arbitrary and continuous resolution adjustments, independently training computationally expensive diffusion models for every specific scale is highly inefficient.

To support various upsampling scales within a single network, arbitrary-scale VSR (AVSR) methods~\cite{chen2022videoinr, chen2023motif, li2024savsr, shang2024stavsr, kim2025bfstvsr, bernasconi2025ldip} have been proposed by adopting coordinate-based implicit neural representations (INRs)~\cite{chen2021liif}. However, since these methods predominantly rely on deterministic regression losses (\eg, MSE, $L_1$), they struggle to capture high-frequency details, often leading to severe over-smoothing at large upscaling factors (\eg, $6\times$ or $8\times$)~\cite{shang2024stavsr, kim2025bfstvsr, kim2024arbitrary}. Motivated by this limitation, recent approaches in the single-image domain~\cite{gao2023idm, kim2024arbitrary, bang2025casarbi} have combined latent diffusion models (LDMs) with INR decoders, demonstrating that generative priors can synthesize high-fidelity details while retaining continuous-scale rendering. Although integrating generative priors with continuous coordinate representations is promising, extending this paradigm to the video domain introduces a distinct set of non-trivial challenges.

\begin{figure}[t]
    \centering
    \includegraphics[width=\linewidth]{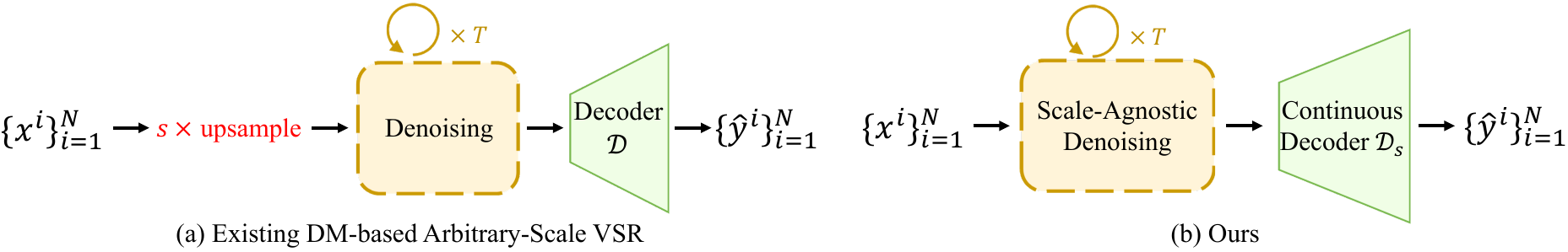}
    \caption{\textbf{Conceptual comparison of DM-based arbitrary-scale VSR.} (a) Existing method~\cite{he2024venhancer} explicitly upsamples the input sequence before the diffusion process. (b) Our AVSR-Diff performs scale-agnostic denoising in the LR latent space, and achieves arbitrary-scale VSR via the continuous video decoder.}
    \label{fig:concept}
    \vspace{-6mm}
\end{figure}

A central hurdle is temporal consistency, as continuous coordinate decoding is highly sensitive to the stability of the underlying features. In frame-wise diffusion models, the inherent stochasticity of per-frame sampling induces minor frame-to-frame variations, which the subsequent decoding process can easily amplify into severe temporal flickering~\cite{yang2024mgldvsr}. While recent DM-based VSR methods introduce various temporal cues, such as warped frame-level guidance~\cite{rota2024stablevsr} and feature-aware modules within the diffusion pipeline~\cite{yang2024mgldvsr, xie2025star, zhou2024upscale, xu2025dgafvsr}, these mechanisms are primarily tailored for \textit{fixed-scale} VSR. Consequently, they do not explicitly provide the scale-agnostic, strictly aligned feature priors required for seamless continuous decoding. Alternatively, relying on full video diffusion models with heavy 3D U-Nets~\cite{he2024venhancer} to explicitly upsample the input sequence prior to denoising, as conceptually illustrated in Fig.~\ref{fig:concept}(a), entangles the heavy denoising process with the target resolution. As empirically demonstrated in Fig.~\ref{fig:teaser}(b), this approach incurs massive memory growth as the scaling factor increases, rendering practical long-video inference fundamentally challenging.

To break this entanglement between the heavy denoising process and target resolution, we propose \textbf{AVSR-Diff} (\textbf{A}rbitrary-scale \textbf{VSR} with \textbf{Diff}usion), a novel decoupled framework designed to ensure both temporal stability and scale-agnostic computational efficiency. As conceptually illustrated in Fig.~\ref{fig:concept}(b), our approach strictly decouples the generative prior extraction from the resolution-specific rendering. We perform diffusion sampling exclusively within a fixed LR latent space, ensuring the generative sampling cost remains completely independent of the target scales (\ie, \textit{scale-agnostic}). However, directly translating these sampled latents via continuous coordinate decoding easily magnifies latent stochasticity into severe temporal flickering. To solve this, we introduce a \textbf{Temporally-Gated Feature Recurrence (TGFR)} module operating within the residual branch of a ControlNet~\cite{zhang2023controlnet}, which guides the frozen denoising U-Net~\cite{rombach2022ldm}. By employing deformable convolutions (DCNs)~\cite{dai2017dcn} for precise sub-pixel feature alignment and an adaptive gate-based fusion strategy, TGFR effectively suppresses drift and flickering artifacts, thereby providing the temporally aligned latent priors required for seamless continuous decoding.

Building upon these temporally aligned priors, we extend the pre-trained fixed-scale image decoder into a \textit{continuous video decoder} to flexibly translate the latent features into arbitrary resolution video frames. To facilitate this transition, we repurpose the TGFR mechanism within the decoding process, ensuring that continuous spatial querying maintains temporal consistency. Crucially, recognizing that the requirement for high-frequency details varies significantly depending on the target scale, we propose a \textbf{Scale-Aware Fourier Refinement (SAFR)} module. Operating within the deep feature space of the frozen VAE decoder, SAFR dynamically modulates the frequency components of the features based on the target scale. These scale-aware features are then continuously queried and decoded by a coordinate-based INR~\cite{chen2021liif}. As a result, this decoupled architecture enables high-fidelity, temporally consistent pixel rendering at any continuous resolution, effectively obviating the need for the computationally expensive HR denoising typically required by video diffusion models.

In summary, the main contributions of our work are three-fold:

\begin{itemize}
    \item We propose AVSR-Diff, a novel decoupled framework for DM-based arbitrary-scale VSR. By separating the diffusion process from the target resolution, our approach performs \textbf{scale-agnostic latent denoising} for computational efficiency, while the proposed \textbf{Temporally-Gated Feature Recurrence (TGFR)} module extracts the temporally aligned priors necessary to support sensitive continuous decoding.

    \item We design an arbitrary-scale \textbf{continuous video decoder} to flexibly translate latent priors into high-fidelity pixels. By integrating a \textbf{Scale-Aware Fourier Refinement (SAFR)} module, our decoder dynamically adapts frequency components to the target scale, effectively synthesizing high-frequency details while rigorously preserving temporal coherence.

    \item Extensive experiments demonstrate the superiority of AVSR-Diff. Remarkably, our arbitrary-scale framework outperforms recent fixed-scale generative models even on their native $4\times$ resolution. Furthermore, it consistently preserves high-frequency details and strong temporal stability across arbitrary scales, surpassing state-of-the-art arbitrary-scale baselines.
\end{itemize}

\section{Related Work}

\subsection{Video Super-Resolution}
Conventional VSR methods~\cite{chan2021basicvsr, chan2022basicvsr++, liang2022rvrt, liang2024vrt, xu2024iart, liu2022ttvsr, tian2020tdan} rely on regression objectives, which inherently produce over-smoothed results. To synthesize high-frequency textures, GAN-based VSR models~\cite{yang2021realvsr, chan2022realbasicvsr, xu2025videogigagan} have been proposed, but they often struggle with a severe consistency-quality dilemma~\cite{xu2025videogigagan}, where their generative capacity amplifies temporal flickering. Recently, driven by the superior perceptual quality of diffusion models (DMs)~\cite{rombach2022ldm, dhariwal2021diffusion}, DM-based VSR approaches have been actively explored. To mitigate the temporal flickering caused by DMs' inherent stochasticity, recent methods incorporate various temporal conditioning mechanisms, such as temporal attention modules, warping-based guidance, and motion-guided sampling, into pre-trained diffusion priors~\cite{rota2024stablevsr, zhou2024upscale, yang2024mgldvsr, xie2025star, xu2025dgafvsr}. Despite their impressive generative capabilities, most mainstream frameworks are restricted to a fixed integer scaling factor (typically $4\times$). While VEnhancer~\cite{he2024venhancer} supports arbitrary-scale upsampling, its reliance on heavy 3D U-Nets for spatio-temporal denoising at the target resolution becomes computationally impractical at large scaling factors. Thus, efficiently bridging generative priors with continuous-scale rendering remains a critical challenge.

\vspace{-3mm}
\subsection{Arbitrary-Scale Super-Resolution}
To overcome the limitations of fixed-scale upsampling, arbitrary-scale super-resolution methods have been proposed by adopting coordinate-based INRs. LIIF~\cite{chen2021liif} pioneered this paradigm in the single-image domain by formulating SR as a continuous image representation. Extending this concept to the video domain, subsequent works~\cite{chen2022videoinr, chen2023motif, li2024savsr, shang2024stavsr, kim2025bfstvsr, bernasconi2025ldip} have actively advanced continuous spatio-temporal representations by incorporating sophisticated motion compensation and temporal feature alignment mechanisms. Despite their architectural flexibility in querying continuous spatial coordinates, these models predominantly rely on deterministic regression objectives during training. Consequently, while they achieve impressive reconstruction metrics, they fundamentally suffer from the perception-distortion trade-off. Particularly at large scaling factors (\eg, $6\times$ or $8\times$), they exhibit severe over-smoothing artifacts and fail to recover the missing high-frequency details.

\vspace{-3mm}
\subsection{Arbitrary-Scale Generative Super-Resolution}
To leverage the powerful generative priors of LDMs for arbitrary-scale rendering, recent studies~\cite{gao2023idm, kim2024arbitrary, bang2025casarbi} have successfully integrated LDMs with coordinate-based INRs. By establishing this continuous generative paradigm in the single-image domain, they can synthesize realistic high-frequency details at any target scale. However, extending this framework to the video domain remains largely unexplored due to a critical bottleneck arising from the incompatibility between the stochastic feature variations of diffusion sampling and the high sensitivity of continuous coordinate decoders, which inevitably amplify temporal flickering. To address this gap, we propose AVSR-Diff. By synergistically integrating the TGFR mechanism for strict temporal feature alignment and the SAFR module for scale-adaptive frequency modulation, our decoupled architecture achieves scale-agnostic computational efficiency. Furthermore, it successfully adapts diffusion priors to arbitrary video scales while rigorously preserving temporal coherence.

\begin{figure}[t]
    \centering
    \includegraphics[width=\textwidth]{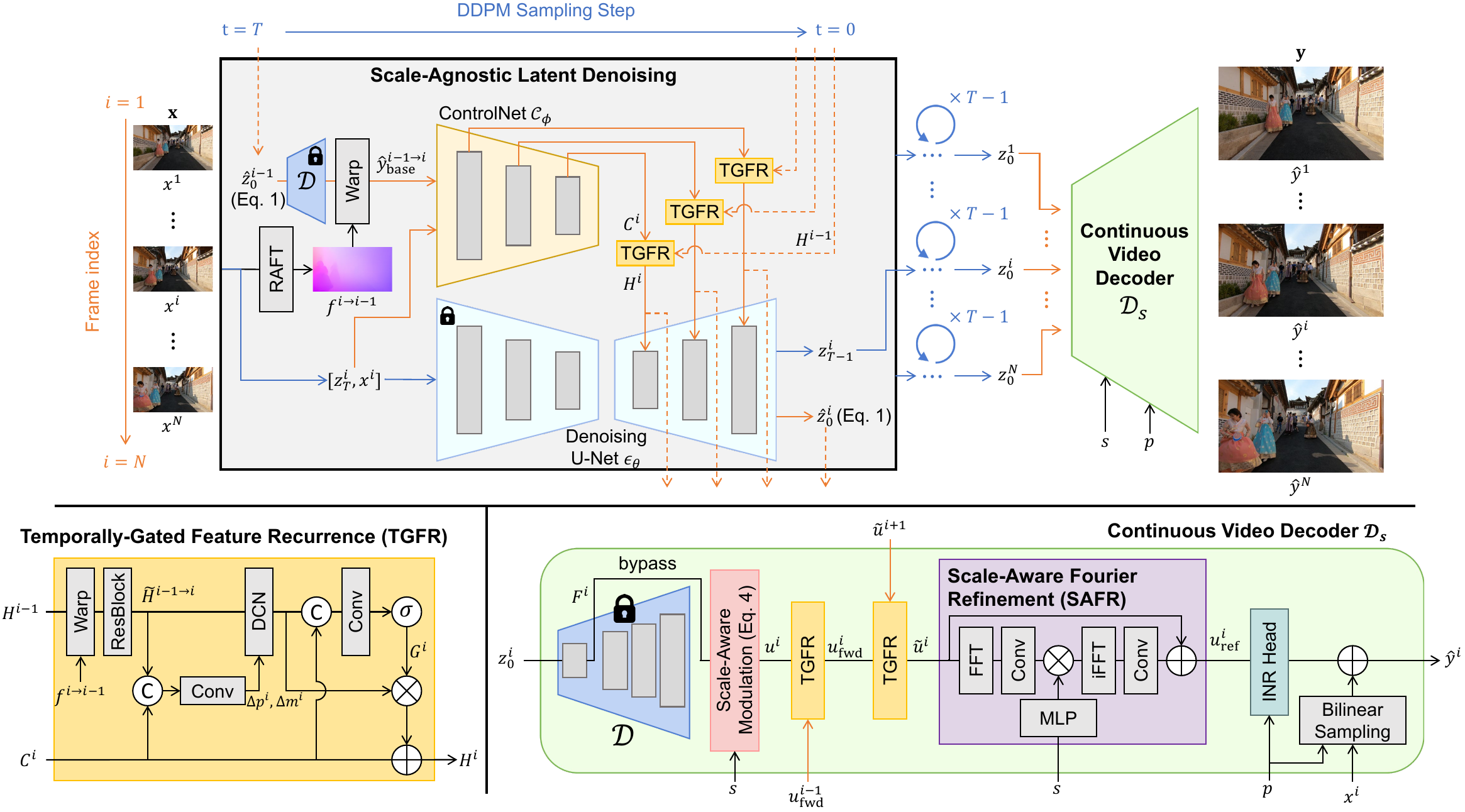}
    \caption{\textbf{Overview of the proposed AVSR-Diff.} A trainable ControlNet ($\mathcal{C}_\phi$) guides the frozen denoising U-Net ($\epsilon_\theta$) for scale-agnostic latent denoising. To enforce temporal consistency, our Temporally-Gated Feature Recurrence (TGFR) module aligns and dynamically gates recurrent features ($H^{i-1}$) across adjacent frames. For arbitrary-scale VSR, the denoised latent sequence ($\mathbf{z}_0 = \{z_0^i\}_{i=1}^N$) is decoded by the Continuous Video Decoder ($\mathcal{D}_s$), which employs the Scale-Aware Fourier Refinement (SAFR) module to conditionally modulate high-frequency details based on the target scale.}
    \label{fig:network}
    \vspace{-3mm}
\end{figure}

\vspace{-3mm}
\section{Method}
\subsection{Overview of AVSR-Diff} \label{sec:overview}
Fig.~\ref{fig:network} shows the overall architectural design of our proposed AVSR-Diff. Given an LR input video sequence $\mathbf{x} = \{x^i\}_{i=1}^N \in \mathbb{R}^{N \times 3 \times H \times W}$ and a target continuous scaling factor $s > 1$, our goal is to reconstruct an HR video sequence $\hat{\mathbf{y}} = \{\hat{y}^i\}_{i=1}^N \in \mathbb{R}^{N \times 3 \times sH \times sW}$ that is both perceptually faithful and temporally consistent. Directly applying diffusion sampling in the HR pixel space is computationally prohibitive and memory-intensive, particularly for video tasks. To overcome this, our AVSR-Diff is designed as a completely decoupled framework built upon a pre-trained single-image super-resolution LDM (SD$\times 4$ Upscaler)~\cite{rombach2022ldm}, consisting of a VAE encoder-decoder pair $(\mathcal{E}, \mathcal{D})$ and a denoising U-Net $\epsilon_\theta$. Our architecture is explicitly decomposed into two stages: scale-agnostic latent denoising and arbitrary-scale continuous decoding.

\noindent\textbf{Scale-Agnostic Latent Denoising.}
To avoid expensive resolution-specific diffusion sampling, as in VEnhancer~\cite{he2024venhancer}, we perform the generative denoising process entirely within a fixed-resolution LR latent space. Since the spatial dimensions of the LR frame $x^i$ naturally match those of the latent space, at each diffusion timestep $t$, we directly concatenate it with the noisy latent $z_t^i \in \mathbb{R}^{C \times H \times W}$ to serve as a structural condition for both the U-Net and the ControlNet~\cite{zhang2023controlnet}. Crucially, to prevent the severe temporal flickering caused by stochastic sampling, we introduce the \textbf{Temporally-Gated Feature Recurrence} (TGFR) module into the ControlNet branch (detailed in Sec.~\ref{sec:tgfr}). The TGFR module explicitly aligns and propagates structural priors across adjacent frames, providing the frozen U-Net $\epsilon_\theta$ with strictly aligned conditions to produce temporally coherent denoising predictions, resulting in a temporally consistent latent sequence $\mathbf{z}_0 = \{z_0^i\}_{i=1}^N \in \mathbb{R}^{N \times C \times H \times W}$.

\noindent\textbf{Continuous Arbitrary-Scale Decoding.}
Given the temporally aligned latent sequence $\mathbf{z}_0$, we directly decode these priors into the target resolution. Specifically, we extend the pre-trained fixed-scale image VAE decoder $\mathcal{D}$ into a continuous video decoder $\mathcal{D}_{s}$ by introducing three key components:
(i) repurposed TGFR blocks for bidirectional feature propagation to maintain temporal coherence within the deep feature space,
(ii) a \textbf{Scale-Aware Fourier Refinement} (SAFR) module that dynamically modulates high-frequency components according to the target scale $s$, and
(iii) a coordinate-based INR~\cite{chen2021liif} for continuous spatial querying.
This decoupled design effectively renders the final HR frames $\hat{\mathbf{y}}$ from the refined features, enabling arbitrary-scale VSR with stable temporal dynamics without the need for resolution-specific diffusion sampling.

\subsection{Temporally-Gated Feature Recurrence (TGFR)} \label{sec:tgfr}
During the scale-agnostic latent denoising process, our goal is to produce temporally coherent latent priors. At each denoising step $t$, the LR frame $x^i$ is directly concatenated with the noisy latent $z_t^i$ to serve as a base structural condition for both the frozen U-Net $\epsilon_\theta$ and the trainable ControlNet $\mathcal{C}_\phi$. While this provides robust spatial guidance for individual frames, stochastic sampling across frames inevitably leads to severe temporal flickering.

To ensure robust temporal consistency, we adopt the temporal conditioning strategy of StableVSR~\cite{rota2024stablevsr}, which encompasses both a frame-wise bidirectional sampling strategy and warped RGB guidance. Since the sampling direction alternates at each diffusion timestep, we formulate the feature propagation from the previous frame $i-1$ to the current frame $i$ within a single forward step for notational simplicity. Under this setting, we first approximate the clean latent $\hat{z}_0^{i-1}$ from $z_t^{i-1}$ using the noise predicted by $\epsilon_\theta$:
\begin{equation}
    \hat{z}_0^{i-1} = \frac{1}{\sqrt{\bar{\alpha}_t}} \left( z_t^{i-1} - \sqrt{1-\bar{\alpha}_t} \epsilon_\theta([z_t^{i-1}, x^{i-1}], t; \mathcal{C}_{\phi}) \right),
\end{equation}
where $\bar{\alpha}_t$ follows the standard noise schedule of the diffusion process. We then decode this latent into the RGB space, $\hat{y}_{\text{base}}^{i-1} = \mathcal{D}(\hat{z}_0^{i-1})$, and backward warp~\cite{wolberg1990digital} it to the current frame using an optical flow $f^{i \rightarrow i-1}$ estimated from the LR inputs by pretrained RAFT~\cite{teed2020raft}. This warped image $\hat{y}_{\text{base}}^{i-1 \rightarrow i} = \mathrm{Warp}(\hat{y}_{\text{base}}^{i-1}, f^{i \rightarrow i-1})$ serves as an explicit pixel-level condition for ControlNet $\mathcal{C}_\phi$. Although $\hat{y}_{\text{base}}^{i-1}$ is decoded at the fixed-scale VAE's native resolution (\eg, $4\times$), it is strictly utilized as a conditioning anchor to stabilize the latent denoising process, and the final arbitrary-scale rendering is handled entirely by our continuous decoder $\mathcal{D}_{s}$.

However, relying solely on warped RGB guidance stabilizes content only at a macroscopic level. Our arbitrary-scale continuous video decoder $\mathcal{D}_{s}$ is highly sensitive to \emph{feature-level} temporal drift, which easily magnifies into flickering artifacts. Therefore, we introduce the \textbf{TGFR} module to recurrently align the deep residual features inside the ControlNet.

Let $C^i$ denote the intermediate residual feature extracted by the ControlNet at the current frame, and $H^{i-1}$ be the recurrent feature propagated from the previous frame. To align $H^{i-1}$ with $C^i$, we first apply a coarse flow-based warping, followed by a refinement block: $\tilde{H}^{i-1 \rightarrow i} = \mathrm{ResBlock}(\mathrm{Warp}(H^{i-1}, f^{i \rightarrow i-1}))$. To rectify the inherent inaccuracies of optical flow at the sub-pixel feature level, we employ a DCN~\cite{dai2017dcn}, inspired by the flow-guided deformable alignment in BasicVSR++~\cite{chan2022basicvsr++}, but repurposed here to recurrently align ControlNet residual features rather than backbone features:
\begin{equation}
    H_{\text{aligned}}^{i-1 \rightarrow i} = \mathrm{DCN}(\tilde{H}^{i-1 \rightarrow i}, \Delta p^i, \Delta m^i),
\label{eq:dcn}
\end{equation}
where the dynamic offsets $\Delta p^i$ and modulation masks $\Delta m^i$ are learned from the concatenated features $[C^i, \tilde{H}^{i-1 \rightarrow i}]$. Finally, while the DCN mask resolves local spatial misalignments, we employ a spatial gated fusion mechanism to evaluate the temporal reliability of the aligned features:
\begin{equation}
\begin{aligned}
    G^i &= \sigma\Big(\mathrm{Conv}\big([C^i, H_{\text{aligned}}^{i-1 \rightarrow i}, |C^i - H_{\text{aligned}}^{i-1 \rightarrow i}|]\big)\Big), \\
    H^i &= C^i + G^i \odot H_{\text{aligned}}^{i-1 \rightarrow i},
\end{aligned}
\label{eq:fusion}
\end{equation}
where $G^i \in [0,1]$ is a learned gate map and $\sigma$ denotes the sigmoid function. By jointly processing the features and their absolute difference $|C^i - H_{\text{aligned}}^{i-1 \rightarrow i}|$, the gate actively suppresses unreliable past information, preventing progressive error accumulation and subsequent temporal artifacts. Note that while the above formulation is defined for a single feature level for brevity, we apply this TGFR mechanism across all multi-scale ControlNet residual feature maps in a recurrent manner. The optical flow $f$ is appropriately resized for each feature scale prior to warping. These fused multi-scale residuals $H^i$ are both propagated to the next frame and injected into the frozen U-Net $\epsilon_\theta$, strictly constraining the denoising trajectory to produce a temporally aligned latent sequence.

\subsection{Arbitrary-Scale Continuous Video Decoding} \label{sec:decoder}

Given the temporally aligned latent sequence $\mathbf{z}_0 = \{z_0^i\}_{i=1}^N$ obtained from the scale-agnostic latent denoising process, our next objective is to render the HR frames $\hat{\mathbf{y}}$ at an arbitrary target scale $s > 1$. Instead of executing computationally heavy diffusion sampling directly at the target resolution, we extend the pre-trained fixed-scale image VAE decoder $\mathcal{D}$ into a continuous video decoder $\mathcal{D}_{s}$ by integrating an implicit coordinate-based renderer and frequency modulation.

\noindent\textbf{Scale-Aware Feature Extraction.}
We bypass the pixel-level reconstruction of the pretrained $\mathcal{D}$, which is strictly bound to a fixed $4\times$ resolution. Instead, given the intermediate deep feature $F^i \in \mathbb{R}^{C' \times H \times W}$ extracted from $\mathcal{D}$ for the latent $z_0^i$, we apply a scale-aware adaptation. Since the optimal high-frequency details required for reconstruction vary depending on the target scale $s$, we modulate the normalized deep features $F^i$ using a sinusoidal scale embedding:
\begin{equation}
    u^i = \mathrm{GN}(F^i) \odot (1 + \gamma(s)) + \beta(s),
\end{equation}
where $\mathrm{GN}$ denotes Group Normalization~\cite{wu2018group}, and $(\gamma(s), \beta(s))$ are channel-wise scale-aware affine parameters predicted by an MLP from the positional encoding~\cite{vaswani2017attention} of $s$.

To ensure that this spatial modulation does not disrupt temporal coherence, and to explicitly extend the single-image representation into a temporal video model, we apply a bidirectional feature propagation mechanism~\cite{schuster1997bidirectional}. By repurposing our TGFR module, we recurrently align and fuse features from adjacent frames in a cascaded manner. Specifically, we sequentially perform forward and backward propagation:
\begin{equation}
\begin{aligned}
    u_{\text{fwd}}^i &= \mathrm{TGFR}(u^i, u_{\text{fwd}}^{i-1}, f^{i \rightarrow i-1}), \\
    \tilde{u}^i &= \mathrm{TGFR}(u_{\text{fwd}}^i, \tilde{u}^{i+1}, f^{i \rightarrow i+1}),
\end{aligned}
\end{equation}
where $\mathrm{TGFR}(v_{\text{curr}}, v_{\text{prior}}, f)$ denotes the DCN-based alignment and gated fusion (as defined in Eqs.~\ref{eq:dcn} and \ref{eq:fusion}) of a prior feature $v_{\text{prior}}$ into the current feature $v_{\text{curr}}$ guided by optical flow $f$. This sequential bidirectional propagation effectively transforms the frame-wise modulated features into the temporally enriched features $\tilde{u}^i$.

\noindent\textbf{Scale-Aware Fourier Refinement (SAFR).}
While the spatial scale conditioning adapts the overall feature amplitude, reconstructing arbitrary-scale textures requires explicit modulation of their frequency-domain representations. As the target scale $s$ increases, high-frequency components need to be dynamically synthesized or preserved to prevent over-smoothing. To achieve this, our SAFR module operates directly in the Fourier domain. Given the temporally enriched feature $\tilde{u}^i$, we compute its frequency representation $U^i = \mathcal{F}(\tilde{u}^i)$. We then apply complex-domain channel mixing followed by a scale-conditioned spectral gate:
\begin{equation}
    u_{\text{ref}}^i = \tilde{u}^i + \mathrm{Conv}\Big(\mathcal{F}^{-1}\big( \mathrm{Mix}_{\mathbb{C}}(U^i) \odot \psi(s) \big)\Big),
\end{equation}
where $\mathcal{F}$ and $\mathcal{F}^{-1}$ denote the 2D Fast Fourier Transform (FFT) and its inverse, respectively, and $\psi(s)$ is a channel-wise spectral gate vector predicted by an MLP. Here, $\mathrm{Mix}_{\mathbb{C}}$ performs $1 \times 1$ convolution in the complex domain to blend channel-wise spectral information, and $\mathrm{Conv}$ is a spatial convolution layer. This spectral gating mechanism explicitly modulates the spectral features conditioned on the target scale $s$.

\noindent\textbf{Continuous Coordinate-Based Rendering.}
Finally, to restore the HR video $\hat{\mathbf{y}}$ without resolution constraints, we employ the LIIF~\cite{chen2021liif} formulation. For any target HR pixel coordinate $p \in [-1, 1]^2$, we query the surrounding local features from the scale-refined map $u_{\text{ref}}^i$ using a continuous MLP to predict the RGB residual $\Delta y^i(p)$. To stabilize the continuous rendering and accelerate convergence, we add a bilinearly sampled color from the original LR input $x^i$:
\begin{equation}
    \hat{y}^i(p) = \Delta y^i(p) + \mathrm{Sample}_{\text{bilinear}}(x^i, p).
\end{equation}
This decoupled architecture effectively renders arbitrary-scale HR frames with strictly preserved temporal dynamics.

\subsection{Training Objectives}
\label{sec:training}

To fully leverage the pre-trained fixed-scale single-image LDM, we freeze the original VAE and the denoising U-Net $\epsilon_\theta$. The ControlNet $\mathcal{C}_\phi$ and the continuous video decoder $\mathcal{D}_s$ are trained independently, as both modules operate within the fixed latent space of the pre-trained LDM.

\noindent\textbf{ControlNet Training.}
We optimize $\mathcal{C}_\phi$ using a combination of a denoising loss and a gate regularization term:
\begin{equation}
    \mathcal{L}_{\text{CNet}} = \| \epsilon - \hat{\epsilon} \|_2^2 + \lambda_{\text{gate}} \frac{1}{|\mathcal{G}|} \sum_{G \in \mathcal{G}} \|G\|_1,
\label{eq:gate}
\end{equation}
where $\epsilon \sim \mathcal{N}(\mathbf{0}, \mathbf{I})$ is added Gaussian noise and $\hat{\epsilon} = \epsilon_\theta([z_t, x], t; \mathcal{C}_\phi)$ is predicted noise. The first term supervises this noise prediction, while the second term applies a sparsity penalty to all spatial gate maps $G$ collected in the set $\mathcal{G}$ from the multi-scale TGFR blocks. This explicitly regularizes the ControlNet $\mathcal{C}_\phi$ to open the gate and propagate features only when the resulting temporal denoising benefit strictly outweighs the activation penalty.

\noindent\textbf{Continuous Decoder Training.}
The arbitrary-scale decoder $\mathcal{D}_s$ is trained in two phases using cropped HR patches $y$ and our predictions $\hat{y}$ at randomly sampled continuous scales $s$:
\begin{equation}
    \mathcal{L}_{\text{Decoder}} = \| \hat{y} - y \|_1 + \lambda_{\text{percep}} \mathcal{L}_{\text{percep}} + \lambda_{\text{adv}} \mathcal{L}_{\text{adv}}.
    \label{eq:decoder}
\end{equation}
In the first phase, $\mathcal{D}_s$ is trained with the $\mathcal{L}_1$ loss and the perceptual loss~\cite{zhang2018unreasonable} $\mathcal{L}_{\text{percep}}$ ($\lambda_{\text{adv}}=0$). In the second phase, we introduce a PatchGAN~\cite{isola2017patch}-based adversarial loss $\mathcal{L}_{\text{adv}}$ to further enhance fine-grained high-frequency realism.

\section{Experiments}

\subsection{Implementation details} \label{sec:implementation}
Our AVSR-Diff is built upon the pre-trained Stable Diffusion $\times 4$ Upscaler (SD$\times 4$ Upscaler)~\cite{rombach2022ldm}. We employ the pre-trained RAFT~\cite{teed2020raft} for optical flow estimation. As described in Sec.~\ref{sec:training}, the ControlNet $\mathcal{C}_\phi$ and the continuous video decoder $\mathcal{D}_s$ are trained independently using the Adam optimizer~\cite{kingma2014adam}. Across all training processes, the batch size is set to 32, and each input sequence consists of 8 LR frames with a spatial size of $64\times64$. For the latent denoising, $\mathcal{C}_\phi$ integrating our TGFR module is trained for 30K steps with a fixed learning rate of 1e-5. During inference, we employ DDPM~\cite{ho2020ddpm} sampling with 50 steps for the denoising process. For the continuous decoding, $\mathcal{D}_s$ is trained in two phases. The first phase optimizes the decoder solely for the $\mathcal{L}_1$ and perceptual losses for 100K steps, using an initial learning rate of 1e-4 that gradually decreases via cosine annealing~\cite{loshchilov2016sgdr}. In the second phase, we introduce the adversarial loss and fine-tune $\mathcal{D}_s$ for an additional 40K steps with a fixed learning rate of 1e-5. During the training of $\mathcal{D}_s$, the target continuous scale $s$ is uniformly sampled from $[1.1, 4.0]$.

\vspace{-2mm}
\subsection{Datasets and Evaluation Metrics}
\noindent\textbf{Datasets.} We train our proposed AVSR-Diff on the standard REDS training set~\cite{nah2019ntire}. For testing, we evaluate on the REDS4~\cite{nah2019ntire} and Vid4~\cite{liu2013bayesian} benchmarks.

\noindent\textbf{Evaluation Metrics.} We comprehensively evaluate the upscaled videos across three distinct aspects. To assess the perceptual quality, we adopt LPIPS~\cite{zhang2018unreasonable} and DISTS~\cite{ding2020image}. For pixel-wise reconstruction fidelity, we report the standard PSNR and SSIM~\cite{wang2004image}. Finally, to rigorously measure temporal consistency and the suppression of flickering artifacts, we employ tLPIPS~\cite{chu2020learning} and tOF~\cite{chu2020learning}. For better readability across all quantitative evaluations, LPIPS, DISTS, and tOF are scaled by $10^2$, and tLPIPS is scaled by $10^3$.

\begin{table*}[t]
\centering
\caption{Quantitative comparison with state-of-the-art methods on the REDS4 and Vid4 datasets for $4\times$ VSR. The best and second-best results among all methods are highlighted in \textbf{\color{red}{red}} and \underline{\color{blue}{blue}}, respectively.}
\label{tab:quantitative_4x}
\resizebox{\linewidth}{!}{
\begin{tabular}{l|cccccc|cccccc}
\toprule
\multirow{2}{*}{Method} & \multicolumn{6}{c|}{REDS4} & \multicolumn{6}{c}{Vid4} \\
\cmidrule(lr){2-7} \cmidrule(lr){8-13}
& LPIPS $\downarrow$ & DISTS $\downarrow$ & PSNR $\uparrow$ & SSIM $\uparrow$ & tLPIPS $\downarrow$ & tOF $\downarrow$ & LPIPS $\downarrow$ & DISTS $\downarrow$ & PSNR $\uparrow$ & SSIM $\uparrow$ & tLPIPS $\downarrow$ & tOF $\downarrow$ \\
\midrule
\multicolumn{13}{c}{Fixed-scale Regression-based VSR} \\
\midrule
BasicVSR++~\cite{chan2022basicvsr++} & 13.49 & 6.99 & \underline{\color{blue}{32.32}} & \underline{\color{blue}{0.9057}} & 9.19 & 18.16 & 19.09 & 12.26 & \underline{\color{blue}{27.72}} & \underline{\color{blue}{0.8409}} & 15.26 & \underline{\color{blue}{4.09}} \\
RVRT~\cite{liang2022rvrt}     & 13.32 & 6.91 & \textbf{\color{red}{32.70}} & \textbf{\color{red}{0.9106}} & 8.98 & 18.08 & 18.81 & 12.09 & \textbf{\color{red}{27.90}} & \textbf{\color{red}{0.8456}} & 15.15 & \textbf{\color{red}{4.06}} \\
\midrule
\multicolumn{13}{c}{Arbitrary-scale Regression-based VSR} \\
\midrule
ST-AVSR~\cite{shang2024stavsr}     & 38.74 & 16.27 & 27.13 & 0.7711 & 24.61 & 23.65 & 41.40 & 18.79 & 24.51 & 0.6863 & 34.47 & 6.72 \\
SAVSR~\cite{li2024savsr}     & 24.30 & 10.15 & 29.41 & 0.8417 & 8.54 & 19.76 & 24.73 & 12.98 & 27.12 & 0.8170 & \underline{\color{blue}{11.83}} & 4.65 \\
\midrule
\multicolumn{13}{c}{Fixed-scale Generative VSR} \\
\midrule
RealBasicVSR~\cite{chan2022realbasicvsr} & 13.40 & 5.99 & 27.04 & 0.7777 & 6.42 & 34.40 & 21.31 & 12.29 & 24.45 & 0.6943 & 36.87 & 7.51 \\
Upscale-A-Video~\cite{zhou2024upscale}    & 40.99 & 16.37 & 24.62 & 0.6495 & 23.68 & 106.60 & 40.90 & 21.38 & 21.88 & 0.5335 & 30.20 & 23.47 \\
MGLD-VSR~\cite{yang2024mgldvsr}       & 14.53 & 6.23 & 26.25 & 0.7408 & 16.36 & 39.62 & 24.74 & 13.97 & 23.51 & 0.6391 & 32.69 & 30.26 \\
StableVSR~\cite{rota2024stablevsr}        & \underline{\color{blue}{9.74}} & \underline{\color{blue}{4.51}} & 27.97 & 0.7951 & \underline{\color{blue}{5.40}} & \underline{\color{blue}{17.20}} & \underline{\color{blue}{18.45}} & \underline{\color{blue}{10.77}} & 24.46 & 0.6989 & 25.26 & 6.09 \\
STAR~\cite{xie2025star}           & 29.48 & 12.17 & 23.08 & 0.6726 & 32.98 & 64.53 & 42.74 & 22.10 & 18.71 & 0.3977 & 46.11 & 22.84 \\
\midrule
\multicolumn{13}{c}{Arbitrary-scale Generative VSR} \\
\midrule
VEnhancer~\cite{he2024venhancer} & 34.69 & 14.91 & 22.90 & 0.6413 & 24.95 & 95.51 & 41.59 & 18.16 & 20.58 & 0.5430 & 29.66 & 14.35 \\
\midrule
\textbf{Ours}  & \textbf{\color{red}{9.54}} & \textbf{\color{red}{4.42}} & 28.75 & 0.8204 & \textbf{\color{red}{4.20}} & \textbf{\color{red}{16.79}} & \textbf{\color{red}{16.79}} & \textbf{\color{red}{9.92}} & 25.12 & 0.7241 & \textbf{\color{red}{9.19}} & 5.78 \\
\bottomrule
\end{tabular}
}
\vspace{-6mm}
\end{table*}

\subsection{Comparison with State-of-the-Art Methods}
We extensively compare our proposed AVSR-Diff against state-of-the-art methods. The baselines comprise fixed-scale regression-based methods (BasicVSR++~\cite{chan2022basicvsr++}, RVRT~\cite{liang2022rvrt}), arbitrary-scale regression-based methods (VideoINR~\cite{chen2022videoinr}, MoTIF~\cite{chen2023motif}, ST-AVSR~\cite{shang2024stavsr}, SAVSR~\cite{li2024savsr}, BF-STVSR~\cite{kim2025bfstvsr}, and V$^3$VSR~\cite{becker2025v3vsr}), fixed-scale generative methods (RealBasicVSR~\cite{chan2022realbasicvsr}, Upscale-A-Video (UAV)~\cite{zhou2024upscale}, MGLD-VSR~\cite{yang2024mgldvsr}, StableVSR~\cite{rota2024stablevsr}, and STAR~\cite{xie2025star}), and the arbitrary-scale generative method VEnhancer~\cite{he2024venhancer}. Table~\ref{tab:quantitative_4x} presents the comparison on the standard $4\times$ VSR, while Table~\ref{tab:quantitative_arb} extends the evaluation to arbitrary continuous scales ($2\times$, $3.25\times$, and $8\times$). Since existing fixed-scale generative methods only support integer scales (\eg, $4\times$), we adapt them for arbitrary scales by first generating the $4\times$ HR videos and subsequently resizing them to the target scale using bicubic interpolation.

\noindent\textbf{Evaluation on Fixed $4\times$ Scale.} 
Table~\ref{tab:quantitative_4x} presents the quantitative comparison on REDS4 and Vid4 for $4\times$ VSR. As expected by the perception-distortion trade-off, regression-based methods achieve higher PSNR but suffer from low perceptual quality. Remarkably, despite being an arbitrary-scale continuous model, our AVSR-Diff outperforms existing fixed-scale generative VSR models on their native $4\times$ resolution. It consistently achieves the best perceptual quality while maintaining the highest fidelity among all generative methods, demonstrating the optimal perception-distortion trade-off. Furthermore, our model exhibits exceptional temporal consistency. It achieves the best tLPIPS across both datasets and the best tOF on REDS4, confirming that our TGFR module successfully suppresses the severe flickering artifacts inherent to diffusion models.

\begin{table*}[t]
\centering
\caption{Quantitative comparison with state-of-the-art methods on the REDS4 dataset for arbitrary-scale VSR ($2\times$, $3.25\times$, and $8\times$).}
\label{tab:quantitative_arb}
\resizebox{\linewidth}{!}{
\begin{tabular}{l|cccc|cccc|cccc}
\toprule
\multirow{2}{*}{Method} & \multicolumn{4}{c|}{$2\times$} & \multicolumn{4}{c|}{$3.25\times$} & \multicolumn{4}{c}{$8\times$} \\
\cmidrule(lr){2-5} \cmidrule(lr){6-9} \cmidrule(lr){10-13}
& LPIPS$\downarrow$ & DISTS$\downarrow$ & PSNR$\uparrow$ & tOF$\downarrow$ & LPIPS$\downarrow$ & DISTS$\downarrow$ & PSNR$\uparrow$ & tOF$\downarrow$ & LPIPS$\downarrow$ & DISTS$\downarrow$ & PSNR$\uparrow$ & tOF$\downarrow$ \\
\midrule
\multicolumn{13}{c}{Arbitrary-scale Regression-based VSR} \\
\midrule
VideoINR~\cite{chen2022videoinr} & 12.26 & 5.49 & 24.87 & 64.41 & 21.96 & 9.11 & 24.24 & 35.42 & 45.31 & 21.31 & 23.31 & 110.73 \\
MoTIF~\cite{chen2023motif}       & 8.39 & 4.08 & 32.36 & 42.46 & 21.38 & 8.20 & 23.02 & 22.03 & 43.63 & 20.72 & 25.36 & 43.26 \\
ST-AVSR~\cite{shang2024stavsr}& 17.42 & 6.43 & 32.51 & 14.04 & 32.55 & 13.26 & \underline{\color{blue}{27.03}} & 21.57 & 59.82 & 27.28 & 24.00 & 43.48 \\
SAVSR~\cite{li2024savsr}                            & 6.66 & 2.95 & 35.25 & \underline{\color{blue}{7.82}} & 19.51 & 7.78 & \textbf{\color{red}{27.13}} & \underline{\color{blue}{18.68}} & 43.02 & 19.53 & \underline{\color{blue}{25.50}} & 46.12 \\
BF-STVSR~\cite{kim2025bfstvsr}                         & 6.43 & 3.15 & 34.74 & 11.17 & 19.85 & 7.89 & 25.55 & 20.13 & 41.02 & 20.19 & 25.38 & \underline{\color{blue}{41.85}} \\
V$^3$VSR~\cite{becker2025v3vsr}              & 6.01 & 2.67 & \textbf{\color{red}{36.13}} & 7.95 & 18.17 & 7.42 & 26.23 & 20.01 & 44.25 & 19.69 & \textbf{\color{red}{25.89}} & 44.51 \\
\midrule
\multicolumn{13}{c}{Fixed-scale Generative VSR + Bicubic} \\
\midrule
RealBasicVSR~\cite{chan2022realbasicvsr} & 8.26 & 4.88 & 31.16 & 22.98 & \underline{\color{blue}{13.03}} & \underline{\color{blue}{5.78}} & 24.47 & 30.34 & 36.00 & 16.18 & 24.14 & 71.04 \\
Upscale-A-Video~\cite{zhou2024upscale}   & 31.27 & 15.03 & 26.32 & 57.35 & 40.78 & 19.94 & 23.47 & 89.66 & 45.46 & 20.11 & 21.70 & 187.20 \\
MGLD-VSR~\cite{yang2024mgldvsr}          & 8.70 & 4.90 & 30.86 & 22.96 & 13.94 & 6.15 & 24.30 & 32.49 & 37.61 & 15.86 & 23.49 & 210.24 \\
StableVSR~\cite{rota2024stablevsr}       & \underline{\color{blue}{5.51}} & \underline{\color{blue}{1.99}} & 33.49 & 8.01 & 14.38 & 6.00 & 24.91 & 20.85 & \underline{\color{blue}{34.96}} & \underline{\color{blue}{15.59}} & 23.72 & 44.09 \\
STAR~\cite{xie2025star}                  & 26.37 & 12.21 & 25.08 & 45.52 & 25.44 & 10.87 & 21.18 & 67.16 & 52.26 & 19.16 & 18.42 & 153.58 \\
\midrule
\multicolumn{13}{c}{Arbitrary-scale Generative VSR} \\
\midrule
VEnhancer~\cite{he2024venhancer} & 23.39 & 10.00 & 23.27 & 77.78 & 30.71 & 12.95 & 22.53 & 83.04 & 43.87 & 19.67 & 21.73 & 129.42 \\
\midrule
\textbf{Ours} & \textbf{\color{red}{3.84}} & \textbf{\color{red}{1.72}} & \underline{\color{blue}{35.47}} & \textbf{\color{red}{7.37}} & \textbf{\color{red}{8.17}} & \textbf{\color{red}{3.21}} & 26.50 & \textbf{\color{red}{15.23}} & \textbf{\color{red}{29.43}} & \textbf{\color{red}{14.09}} & 24.95 & \textbf{\color{red}{39.13}} \\
\bottomrule
\end{tabular}
}
\vspace{-4mm}
\end{table*}

\noindent\textbf{Evaluation on Arbitrary Scales.} 
Table~\ref{tab:quantitative_arb} demonstrates our model's superiority across continuous ($3.25\times$) and large ($8\times$) scales. While fixed-scale generative models suffer from severe aliasing when relying on bicubic interpolation, AVSR-Diff directly renders HR frames, consistently outperforming them and the recent arbitrary-scale VEnhancer~\cite{he2024venhancer} in both perceptual quality and temporal consistency. Notably, at the $2\times$ scale, AVSR-Diff achieves a PSNR (35.47 dB) comparable to strong regression-based baselines despite being a generative model, while delivering far superior perceptual quality and temporal stability, effectively bridging the perception-distortion gap. This confirms our continuous video decoder's robustness and fine-detail preservation for arbitrary and large-scale VSR.

\begin{figure*}[t]
\centering
\includegraphics[width=\linewidth]{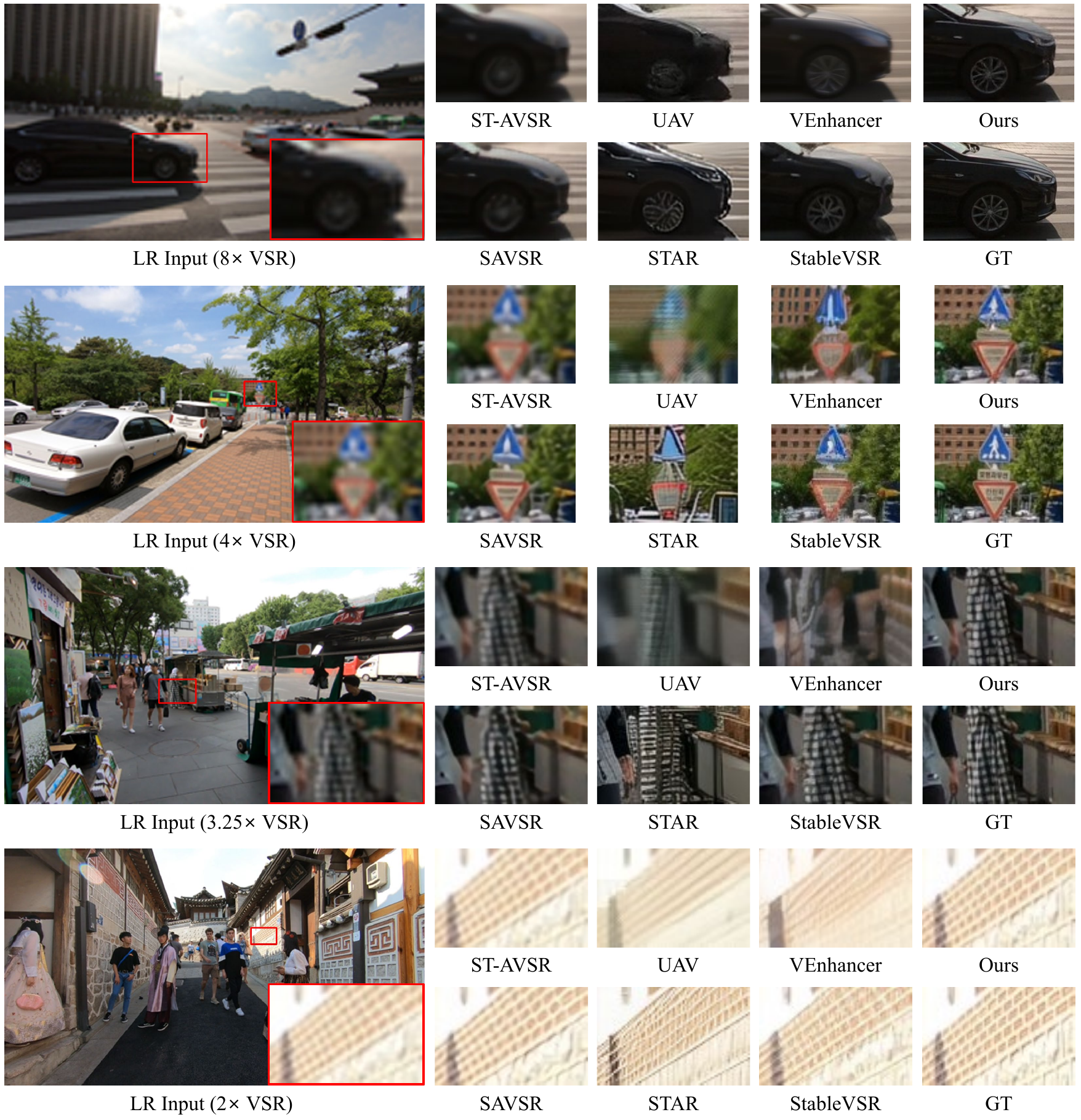}
\vspace{-6mm}
\caption{Qualitative comparison across various upscaling factors on REDS4 dataset~\cite{nah2019ntire}.}
\label{fig:qualitative}
\vspace{-6mm}
\end{figure*}

\noindent\textbf{Qualitative Results.}
Fig.~\ref{fig:qualitative} and Fig.~\ref{fig:teaser}(a) provide visual comparisons across various scales. As observed, regression-based methods inherently produce overly smoothed outputs, failing to recover complex textures. Conversely, while recent generative models actively synthesize high-frequency details, they frequently fail to preserve the original structural fidelity, often generating incorrect patterns or structurally distorted artifacts. In contrast, our AVSR-Diff consistently maintains the structural identity of the LR inputs. By effectively balancing robust reconstruction fidelity with generative priors, our method synthesizes sharp, fine-grained textures that remain faithful to the original content, without introducing unnatural structural distortions, even at large scaling factors (\eg, $8\times$). Further quantitative and qualitative comparisons across diverse scales are provided in the \textit{Suppl}.

\vspace{-2mm}
\subsection{Ablation Study} \label{sec:ablation}
\noindent\textbf{Effectiveness of Key Components.}
Table~\ref{tab:ablation} details the progressive improvements achieved by our proposed modules. Starting from the baseline (a), which represents StableVSR~\cite{rota2024stablevsr}, directly substituting the pretrained decoder $\mathcal{D}$ with our continuous video decoder $\mathcal{D}_s$ in (b) fails to stabilize the decoding process and even degrades the temporal metrics. This directly validates that continuous coordinate rendering is sensitive to latent stochasticity, making explicit feature-level alignment essential. To resolve this, we introduce the TGFR module. Integrating flow-based coarse alignment (c) and DCN-based sub-pixel refinement (d) strictly aligns the temporal priors and steadily improves both temporal consistency and reconstruction fidelity. With these aligned priors, re-integrating our continuous video decoder in (e) not only preserves but further enhances temporal stability. However, this transition inherently compromises fine-grained details, as evidenced by the simultaneous degradation in perceptual metrics. Remarkably, the integration of our SAFR module (Ours) effectively recovers these high-frequency components, yielding the best overall perceptual quality and temporal consistency.

\begin{table}[t]
\centering
\caption{Ablation study on the key components of AVSR-Diff. The evaluation is conducted on the REDS4 dataset for $4\times$ VSR. `Flow' and `DCN' denote the alignment strategies within the TGFR module. $\mathcal{D}$ and $\mathcal{D}_s$ represent the pre-trained single-image decoder and our continuous video decoder, respectively.}
\label{tab:ablation}
\resizebox{10cm}{!}{
\begin{tabular}{c|cc|ccc|cccccc}
\toprule
\multirow{2}{*}{Model\hspace{2mm}} & \multicolumn{2}{c|}{TGFR} & \multicolumn{3}{c|}{Decoder} & \multirow{2}{*}{\hspace{2mm}LPIPS$\downarrow$} & \multirow{2}{*}{DISTS$\downarrow$} & \multirow{2}{*}{PSNR$\uparrow$} & \multirow{2}{*}{SSIM$\uparrow$} & \multirow{2}{*}{tLPIPS$\downarrow$} & \multirow{2}{*}{tOF$\downarrow$\hspace{2mm}} \\
\cmidrule(lr){2-3} \cmidrule(lr){4-6}
& \hspace{2mm}Flow\hspace{2mm} & DCN\hspace{2mm} & \hspace{2mm}$\mathcal{D}$\hspace{2mm} & \hspace{1.5mm}$\mathcal{D}_s$\hspace{1.5mm} & SAFR\hspace{2mm} & & & & & & \\
\midrule
(a) & & & \checkmark & & & 9.74 & 4.51 & 27.97 & 0.7951 & 5.40 & 17.20 \\ 
(b) & & & & \checkmark & \checkmark & 9.65 & 4.48 & 28.30 & 0.8075 & 5.45 & 17.24 \\
(c) & \checkmark & & \checkmark & & & 9.73 & 4.45 & 28.24 & 0.8043 & 5.12 & 17.17 \\ 
(d) & \checkmark & \checkmark & \checkmark & & & 9.71 & 4.52 & 28.35 & 0.8099 & 4.95 & 17.05 \\ 
(e) & \checkmark & \checkmark & & \checkmark & & 10.06 & 4.59 & 28.61 & 0.8141 & 4.43 & 16.92 \\ 
\midrule
\textbf{Ours} & \checkmark & \checkmark & & \checkmark & \checkmark & \textbf{9.54} & \textbf{4.42} & \textbf{28.75} & \textbf{0.8204} & \textbf{4.20} & \textbf{16.79} \\
\bottomrule
\end{tabular}
}
\vspace{-4mm}
\end{table}

\noindent\textbf{Long-term Stability and Gate Design.}
Recurrently propagating ControlNet residual features can accumulate errors over long sequences. Table~\ref{tab:gate_ablation} isolates the effects of the learned gate and its sparsity regularization by comparing clip-wise and single-pass inference on the same $100$-frame sequences. When processed as five consecutive $20$-frame clips, both ablated variants complete inference without collapse, yet underperform the full TGFR. In a single $100$-frame pass, however, replacing gated fusion with concatenation after the same flow-guided DCN alignment (w/o Gate (Concat.)) leads to structural collapse, indicating that motion alignment alone cannot prevent unreliable recurrent information from accumulating over time. 

\begin{figure}[b]
    \vspace{-4mm}
    \centering
    \includegraphics[width=\linewidth]{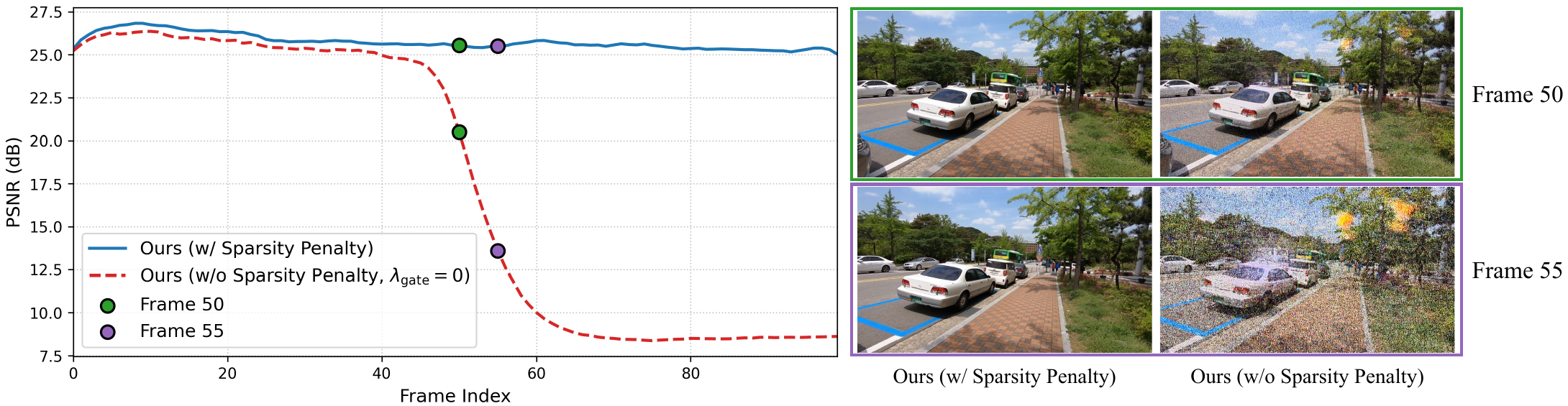}
    \vspace{-5mm}
    \caption{Effect of the gate sparsity penalty on long-term temporal stability. Without it, progressive error accumulation causes severe structural noise at later frames.}
    \label{fig:long_term}
\end{figure}

Retaining the learned gate but removing its sparsity penalty (Gate w/o Sparsity, $\lambda_{\text{gate}}=0$) is likewise unstable, degenerating after roughly $50$ frames as visualized in Fig.~\ref{fig:long_term}. By discouraging unnecessary gate activations and promoting selective temporal propagation, the sparsity term enables the full TGFR to remain stable and achieve the best performance across all metrics throughout the single $100$-frame pass.

\begin{figure}[h]
    \vspace{-3mm}
    \centering
    \includegraphics[width=\linewidth]{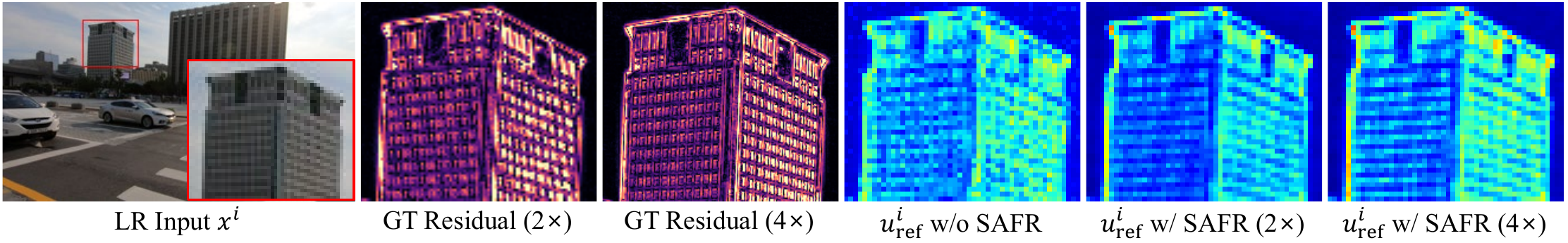}
    \vspace{-5mm}
    \caption{Visualization of scale-aware feature representations. Compared to a static baseline (w/o SAFR module), our SAFR module dynamically adapts feature activations ($u_{\text{ref}}^i$) to match the target high-frequency residuals.}
    \label{fig:safr_vis}
    \vspace{-5mm}
\end{figure}

\noindent\textbf{Visualization of Scale-Aware Refinement.}
To deeply understand how our continuous video decoder $\mathcal{D}_s$ adapts to arbitrary resolutions, we visualize the scale-aware feature representations $u_{\text{ref}}^i$ before the coordinate-based rendering. Fig.~\ref{fig:safr_vis} compares the feature activation heatmaps alongside the target high-frequency residuals (GT - Bilinear) for both $2\times$ and $4\times$ upscaling factors. Without our scale-aware modules, the network produces a static, scale-agnostic feature map that fails to accommodate the varying high-frequency demands of different target resolutions. This rigidity leads to over-smoothing at large scales and aliasing at small scales. In contrast, our SAFR module dynamically modulates the spectral energy based on the target scale $s$. At the $2\times$ scale, the feature activation is tightly concentrated on structural edges to prevent over-enhancement. Conversely, at the $4\times$ scale, SAFR explicitly amplifies the high-frequency activations to synthesize the missing complex textures. This scale-adaptive behavior correlates with the target residual distributions, confirming that our scale-aware refinement successfully bridges generative priors with arbitrary-scale continuous decoding.

\begin{table}[t]
\centering
\caption{Ablation of the TGFR gate design under long-sequence inference on REDS4 ($4\times$). Each $100$-frame sequence is processed either as five consecutive $20$-frame clips (recurrence reset at boundaries) or a single $100$-frame pass. w/o Gate (Concat.) replaces gated fusion with concatenation, and Gate w/o Sparsity removes the sparsity penalty.}\label{tab:gate_ablation}
\resizebox{0.8\linewidth}{!}{
\begin{tabular}{l|cccc|cccc}
\toprule
\multirow{2}{*}{Variant} & \multicolumn{4}{c|}{Five consecutive $20$-frame clips} & \multicolumn{4}{c}{Single $100$-frame pass} \\
\cmidrule(lr){2-5} \cmidrule(lr){6-9}
& LPIPS$\downarrow$ & PSNR$\uparrow$ & tLPIPS$\downarrow$ & tOF$\downarrow$ & LPIPS$\downarrow$ & PSNR$\uparrow$ & tLPIPS$\downarrow$ & tOF$\downarrow$ \\
\midrule
w/o Gate (Concat.)   & 12.59 & 27.39 & 25.48 & 27.28 & 42.65 & 19.53 & 91.22 & 1887.47 \\
Gate w/o Sparsity    & 11.85 & 27.61 & 22.74 & 25.85 & 40.57 & 20.06 & 88.87 & 1718.06 \\
\textbf{Full TGFR}   & \textbf{11.05} & \textbf{28.42} & \textbf{7.52} & \textbf{18.52} & \textbf{9.54} & \textbf{28.75} & \textbf{4.20} & \textbf{16.79} \\
\bottomrule
\end{tabular}
}
\vspace{-4mm}
\end{table}

\vspace{-3mm}
\section{Conclusion}
\vspace{-1mm}
We present AVSR-Diff, a novel decoupled framework for DM-based arbitrary-scale VSR. By separating scale-agnostic latent denoising from continuous arbitrary-scale decoding, AVSR-Diff successfully bridges the gap between reconstruction fidelity and perceptual quality without incurring heavy resolution-specific sampling costs, making it highly memory-efficient. Within this framework, the Temporally-Gated Feature Recurrence (TGFR) module robustly suppresses diffusion-inherent flickering, while the Scale-Aware Fourier Refinement (SAFR) integrated into the continuous video decoder $\mathcal{D}_s$ synthesizes fine-grained high-frequency details across arbitrary resolutions. Furthermore, the introduction of an explicit gate sparsity penalty effectively prevents progressive error accumulation, ensuring exceptional stability in long-term sequence processing. Extensive evaluations on standard benchmarks demonstrate that AVSR-Diff not only outperforms existing continuous models but also surpasses recent fixed-scale generative methods on their native $4\times$ resolution, establishing a new optimal perception-distortion trade-off in the VSR domain.
\end{sloppypar}

\subsubsection*{Acknowledgements.}
This work was supported by Institute of Information \& communications Technology Planning \& Evaluation (IITP) grant funded by the Korean Government [Ministry of Science and ICT (Information and Communications Technology)] (Project Number: RS-2022-00144444, Project Title: Deep Learning Based Visual Representational Learning and Rendering of Static and Dynamic Scenes).


%
%

\clearpage
\bibliographystyle{splncs04}
\bibliography{main}

\clearpage
\renewcommand{\thefigure}{S\arabic{figure}}
\renewcommand{\thetable}{S\arabic{table}}
\renewcommand{\theequation}{S\arabic{equation}}
\renewcommand{\thealgorithm}{S\arabic{algorithm}}
\renewcommand{\thesection}{S\arabic{section}}
\renewcommand{\thesubsection}{S\arabic{section}.\arabic{subsection}}
\setcounter{figure}{0}
\setcounter{table}{0}
\setcounter{equation}{0}
\setcounter{section}{0}

\title{AVSR-Diff: Supplementary Material}
\author{}
\institute{}
\maketitle
In this \textit{Supplementary Material}, we provide additional details and results to support the main paper. First, in Sec.~\ref{sec:add_implementation}, we provide the additional implementation details, including the loss weights and hardware environment. Second, in Sec.~\ref{sec:inference_strategy}, we elaborate on our bounding strategy for out-of-distribution scales. Third, in Sec.~\ref{sec:pipeline}, we present the complete algorithmic representation of our proposed decoupled inference pipeline. Furthermore, we report additional quantitative results in Sec.~\ref{sec:supp_quant} and additional qualitative comparisons in Sec.~\ref{sec:supp_qual}. Finally, we analyze the scale-wise computational efficiency of AVSR-Diff in Sec.~\ref{sec:efficiency}, and discuss its limitations and future directions in Sec.~\ref{sec:supp_limit}.

\section{Additional Implementation Details} \label{sec:add_implementation}

The training of our AVSR-Diff is conducted on two NVIDIA RTX A6000 GPUs. During the training of the ControlNet $\mathcal{C}_\phi$, the weight for the explicit gate sparsity penalty (Eq.~\ref{eq:gate} of the main paper) is set to $\lambda_{\text{gate}} = 0.01$. For the training of the continuous video decoder $\mathcal{D}_s$, we empirically set the perceptual loss weight (Eq.~\ref{eq:decoder} of the main paper) to $\lambda_{\text{percep}} = 1.0$ across all phases, and introduce the adversarial loss (Eq.~\ref{eq:decoder} of the main paper) with a weight of $\lambda_{\text{adv}} = 0.05$ specifically during the second fine-tuning phase. Furthermore, during the continuous coordinate-based rendering phase, to maintain a constant memory footprint regardless of the target scale $s$, we evaluate the queried spatial coordinates in chunks (\eg, 50,000 points per batch). This memory-efficient implementation explicitly bounds the peak GPU memory, ensuring it remains invariant to the output resolution, as demonstrated in Fig.~\ref{fig:teaser}(b) of the main paper.

\begin{figure}[t]
    \centering
    \includegraphics[width=\linewidth]{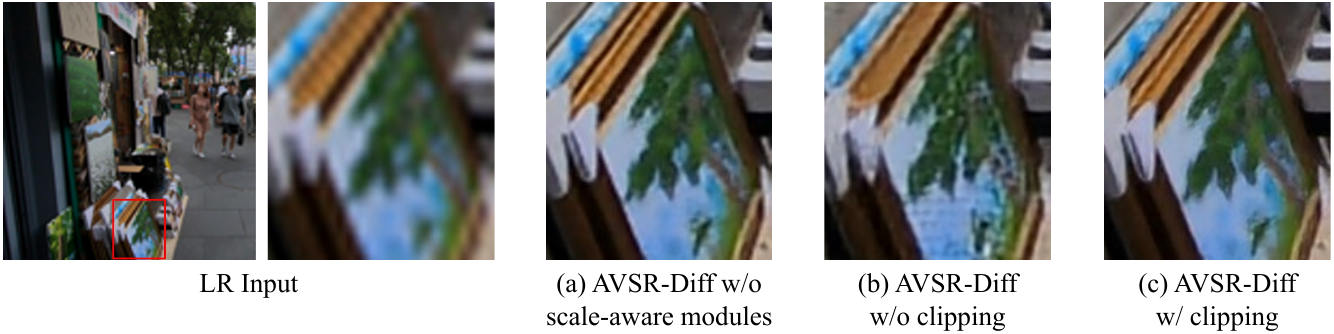} 
    \vspace{-5mm}
    \caption{Visual comparison of scale conditioning strategies at a large out-of-distribution scale ($8\times$). (a) The baseline \textbf{without scale-aware modules} exhibits relatively softer textures and lacks some fine-grained high-frequency details. (b) \textbf{AVSR-Diff without clipping} ($s=8$) suffers from severe spatial artifacts due to positional embedding extrapolation. (c) \textbf{AVSR-Diff with clipping} ($s=4$) successfully balances robust generative priors and structural stability. It suppresses the severe OOD artifacts while synthesizing sharper and more faithful textures than the baseline (a).}
    \label{fig:supp_ood}
    \vspace{-5mm}
\end{figure}

\begin{table}[h]
\centering
\vspace{-3mm}
\caption{Quantitative comparison of scale conditioning strategies on the REDS4 dataset at a large out-of-distribution scale ($8\times$). For better readability, LPIPS, DISTS, and tOF are scaled by $10^2$, and tLPIPS is scaled by $10^3$.}
\vspace{-1mm}
\label{tab:supp_ood_quantitative}
\resizebox{\textwidth}{!}{
\begin{tabular}{l cccccc}
\hline
\textbf{Model / Strategy} & \textbf{LPIPS$\downarrow$} & \textbf{DISTS$\downarrow$} & \textbf{PSNR$\uparrow$} & \textbf{SSIM$\uparrow$} & \textbf{tLPIPS$\downarrow$} & \textbf{tOF$\downarrow$} \\
\hline
(a) AVSR-Diff w/o Scale-aware modules & 31.01 & 15.10 & 24.90 & 0.6648 & 6.64 & 41.11 \\
(b) AVSR-Diff w/o clipping ($s=8$) & 36.02 & 15.23 & 24.78 & 0.6594 & 9.92 & 46.43 \\
(c) AVSR-Diff w/ clipping ($s=4$) & \textbf{29.43} & \textbf{14.09} & \textbf{24.95} & \textbf{0.6653} & \textbf{6.39} & \textbf{39.13} \\
\hline
\end{tabular}
}
\vspace{-6mm}
\end{table}

\vspace{-3mm}
\section{Inference Strategy for Out-of-Distribution Scales} \label{sec:inference_strategy}

While AVSR-Diff is trained on continuous scale factors up to $4\times$, we evaluate its robustness on large out-of-distribution (OOD) scaling factors (\eg, $8\times$). Although our continuous implicit neural representation (INR) natively supports arbitrary resolutions, directly feeding an OOD scale factor into the scale-aware modules (Scale-Aware Modulation and SAFR) causes spatial artifacts due to the extrapolation limits of absolute positional encodings (see Fig.~\ref{fig:supp_ood}).

To mitigate this, we adopt a simple bounding strategy during large-scale inference. The scale condition injected into the scale-aware modules is explicitly clipped to $s=4$ (the maximum training scale), while the subsequent INR still queries the exact continuous coordinates for the target resolution. As shown in Table~\ref{tab:supp_ood_quantitative} and Fig.~\ref{fig:supp_ood}, this clipping strategy successfully suppresses OOD artifacts while preserving the fine-grained generative priors learned at the $4\times$ boundary, outperforming both the unclipped setting and the baseline without scale-aware conditioning.

\vspace{-3mm}
\section{Complete Inference Pipeline} \label{sec:pipeline}
To provide a comprehensive and transparent overview of our proposed method, we detail the complete inference pipeline of AVSR-Diff. To explicitly demonstrate how the framework is decoupled into two distinct stages, we present the pseudo-code in two separate algorithms. Specifically, Algorithm~\ref{alg:stage1} outlines the frame-wise bidirectional sampling process for scale-agnostic latent denoising. Subsequently, Algorithm~\ref{alg:stage2} details the sequence-level continuous arbitrary-scale decoding.

\begin{figure}[t]
    \centering
    \includegraphics[width=\linewidth]{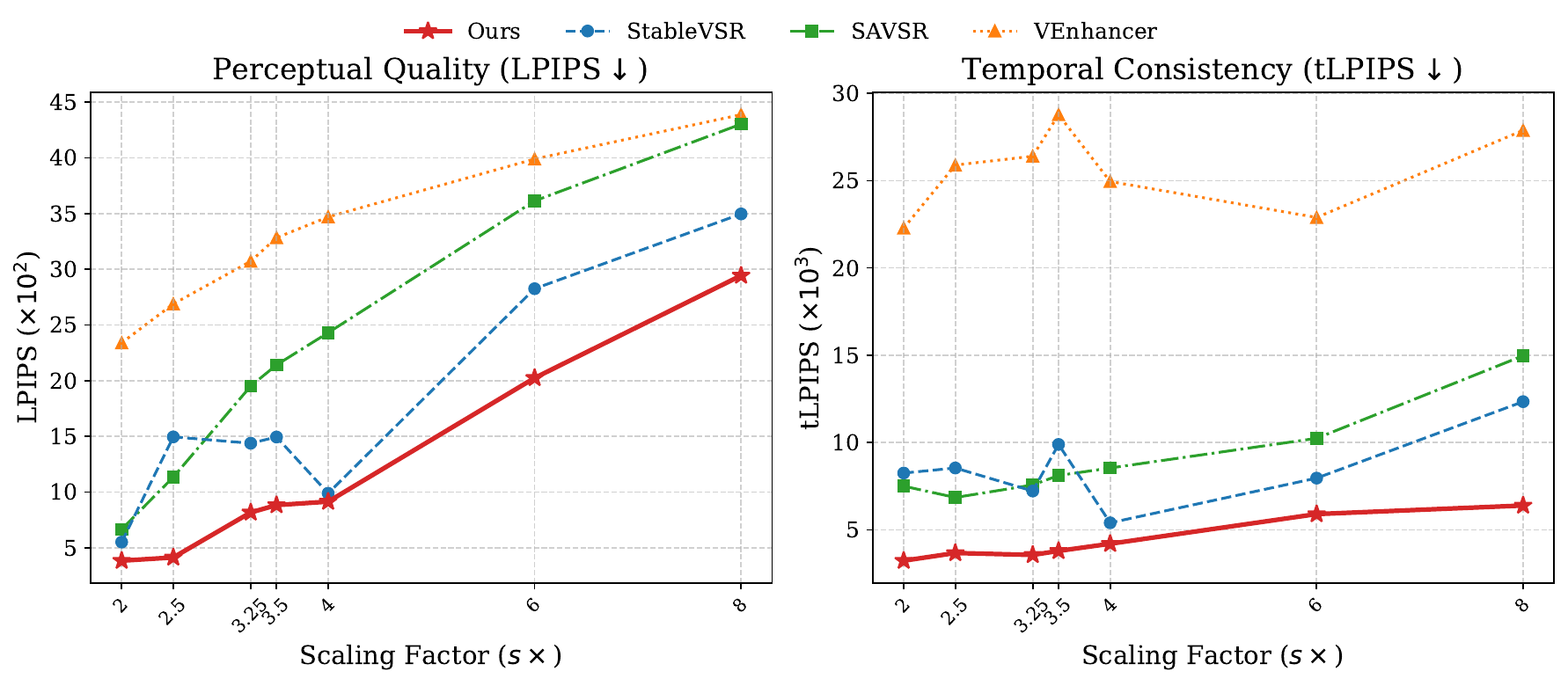} 
    \vspace{-6mm}
    \caption{\textbf{Quantitative performance trends across continuous scaling factors (from $2\times$ to $8\times$).} We plot the LPIPS (left) and tLPIPS (right) scores of our AVSR-Diff against recent state-of-the-art baselines. The plotted values correspond to the quantitative results reported in Table~\ref{tab:supp_master_table} and Table~\ref{tab:quantitative_4x} of the main paper. As the target scaling factor increases, existing methods suffer from perceptual degradation and temporal flickering. In contrast, our method consistently maintains the lowest LPIPS and tLPIPS scores across all evaluated scales, explicitly demonstrating its robust arbitrary-scale generative capability and temporal consistency.}
    \label{fig:supp_trend}
    \vspace{-3mm}
\end{figure}

\vspace{-2mm}
\section{Additional Quantitative Results} \label{sec:supp_quant}
\vspace{-2mm}
Due to space constraints, Table~\ref{tab:quantitative_arb} of the main paper reports only a subset of metrics (LPIPS, DISTS, PSNR, and tOF) at $2\times$, $3.25\times$, and $8\times$. Here, Table~\ref{tab:supp_master_table} extends the evaluation to additional scales ($2.5\times$, $3.5\times$, and $6\times$) and reports all six metrics, including SSIM and tLPIPS, at each reported scale.

As shown in Table~\ref{tab:supp_master_table} and Fig.~\ref{fig:supp_trend}, AVSR-Diff performs consistently well across diverse upscaling scenarios. At $2\times$, it achieves the best perceptual quality (LPIPS and DISTS) and temporal stability (tLPIPS and tOF) among the evaluated methods, while remaining competitive with the strongest regression-based baselines in pixel-level fidelity (PSNR and SSIM). At the remaining arbitrary and out-of-distribution scales, AVSR-Diff consistently leads in perceptual quality and temporal stability. These results support the effectiveness of our decoupled framework in maintaining strong perceptual quality and temporal consistency across a wide range of target scales, including scales beyond the training range.

\vspace{-2mm}
\section{Additional Qualitative Results} \label{sec:supp_qual}
\vspace{-2mm}

\begin{figure}[t]
    \centering
    \includegraphics[width=\linewidth]{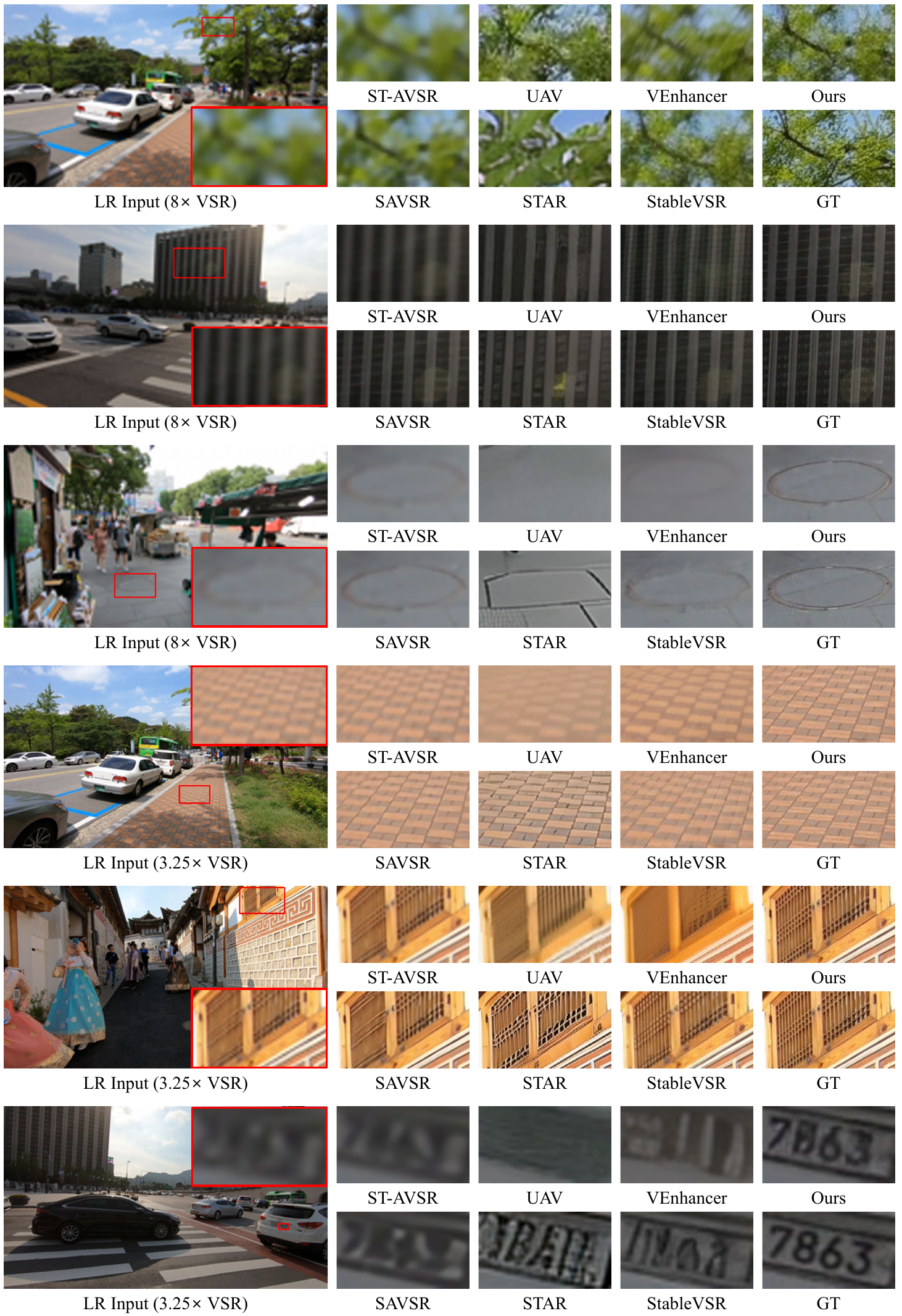} 
    \vspace{-6mm}
    \caption{\textbf{Additional qualitative comparisons at arbitrary ($3.25\times$) and large ($8\times$) scaling factors.} Compared to recent regression-based and DM-based methods, our AVSR-Diff achieves an optimal perceptual-distortion trade-off. It successfully synthesizes highly-detailed textures while preserving the structural fidelity of the original LR inputs without introducing unnatural artifacts, even at large scaling factors.}
    \label{fig:supp_qual}
\end{figure}

To further validate our proposed AVSR-Diff, we provide extensive visual comparisons at both arbitrary ($3.25\times$) and large ($8\times$) scaling factors, as shown in Fig. \ref{fig:supp_qual}. Consistent with the main paper, regression-based methods inherently produce overly smoothed outputs and struggle to recover complex textures. Furthermore, while recent DM-based models actively synthesize fine-grained details, they frequently fail to preserve the structural fidelity of the original LR inputs, leading to incorrect patterns or severe structural distortions, especially at large scaling factors. In contrast, our AVSR-Diff successfully achieves an optimal perceptual-distortion trade-off. By effectively balancing robust reconstruction fidelity with generative priors, our method synthesizes sharp, photo-realistic textures that remain faithful to the original content without introducing unnatural artifacts, even at the large $8\times$ scale.

\vspace{-2mm}
\section{Scale-wise Efficiency Analysis} \label{sec:efficiency}
\vspace{-2mm}
Table~\ref{tab:efficiency} compares AVSR-Diff with VEnhancer~\cite{he2024venhancer}, a directly comparable arbitrary-scale video diffusion baseline, on the same 10-frame $180\times320$ LR sequence. AVSR-Diff performs denoising in a fixed-resolution LR latent space, maintaining a peak memory of $8.7$\,GB and a runtime of approximately $42$\,s/frame across $2\times$ to $8\times$. In contrast, VEnhancer performs denoising at the target resolution, causing its peak memory and runtime to increase substantially with the target scale, from $40.3$\,GB to $72.4$\,GB and from $24.88$ to $106.52$\,s/frame. For AVSR-Diff, the chunked INR querying only batches independent output-coordinate queries in the final rendering stage and does not partition the diffusion denoising process. Although VEnhancer requires fewer absolute FLOPs and is faster at $2\times$ and $4\times$, its cost grows rapidly with the target scale: from $2\times$ to $8\times$, the total FLOPs of AVSR-Diff increase by only $1.35\%$, whereas those of VEnhancer increase by $13.5\times$. Consequently, AVSR-Diff becomes approximately $2.5\times$ faster than VEnhancer at $8\times$, while requiring substantially less peak memory. Overall, these results demonstrate that the efficiency advantage of AVSR-Diff is \emph{scale-wise} rather than absolute: its memory footprint and computational cost remain nearly invariant as the requested output resolution increases.

\begin{table}[t]
\centering
\caption{Scale-wise efficiency comparison with VEnhancer on the same 10-frame $180\times320$ LR sequence. We report peak GPU memory, runtime per frame, and TFLOPs per frame at different target scales. VEnhancer uses tiled inference. AVSR-Diff has higher absolute FLOPs due to 50-step sampling, but its memory and computation remain nearly constant as the target scale increases.}
\vspace{-3mm}
\label{tab:efficiency}
\resizebox{\linewidth}{!}{
\begin{tabular}{l|ccc|ccc|ccc|ccc}
\toprule
\multirow{2}{*}{Method} & \multicolumn{3}{c|}{$2\times$} & \multicolumn{3}{c|}{$4\times$} & \multicolumn{3}{c|}{$8\times$} & \multicolumn{3}{c}{Growth ($2\times\!\rightarrow\!8\times$)} \\
\cmidrule(lr){2-4} \cmidrule(lr){5-7} \cmidrule(lr){8-10} \cmidrule(lr){11-13}
& Mem.$\downarrow$ & Time$\downarrow$ & FLOPs$\downarrow$ & Mem.$\downarrow$ & Time$\downarrow$ & FLOPs$\downarrow$ & Mem.$\downarrow$ & Time$\downarrow$ & FLOPs$\downarrow$ & \multirow{2}{*}{Mem.} & \multirow{2}{*}{Time} & \multirow{2}{*}{FLOPs} \\
& (GB) & (s/fr.) & (T/fr.) & (GB) & (s/fr.) & (T/fr.) & (GB) & (s/fr.) & (T/fr.) & & & \\
\midrule
VEnhancer~\cite{he2024venhancer} (tiled) & 40.3 & 24.88 & 15.10 & 49.2 & 35.89 & 58.38 & 72.4 & 106.52 & 204.52 & 1.8$\times$ & 4.3$\times$ & 13.5$\times$ \\
AVSR-Diff & 8.7 & 41.87 & 219.44 & 8.7 & 41.98 & 220.04 & 8.7 & 42.27 & 222.41 & \textbf{1.0$\times$} & \textbf{1.01$\times$} & \textbf{1.01$\times$} \\
\bottomrule
\end{tabular}
}
\vspace{-5mm}
\end{table}

\vspace{-2mm}
\section{Limitations and Future Work} \label{sec:supp_limit}
\vspace{-2mm}

\noindent\textbf{Absolute inference cost.}
Despite the favorable scale-wise efficiency demonstrated in Sec.~\ref{sec:efficiency}, AVSR-Diff remains computationally expensive in absolute terms due to its iterative diffusion sampling. Unlike INR-based models that synthesize outputs in a single forward pass, our method relies on a diffusion-based generative prior involving 50 sampling steps, as detailed in Sec.~\ref{sec:implementation} of the main paper. For instance, upsampling a $180\times320$ input frame takes approximately 42 seconds. Accelerating this sampling process is a valuable direction for future research; we believe that integrating advanced sampling strategies~\cite{lu2022dpm, karras2022elucidating} or recent diffusion distillation techniques~\cite{lin2025diffusion, zhang2024sf, luo2023latent} could substantially reduce the inference time without compromising the structural fidelity and temporal coherence achieved by our model.

\noindent\textbf{Fixed-scale generative prior.}
A further limitation stems from our reliance on the pretrained SD$\times$4 Upscaler~\cite{rombach2022ldm}: since the generative prior is bounded at a fixed $4\times$ scale, at larger target scales (\eg, $8\times$) the continuous decoder extrapolates in feature space rather than drawing on additional generative capacity. We mitigate this via the scale-clipping strategy (Sec.~\ref{sec:inference_strategy}), but exploring generative backbones trained across a range of native resolutions remains a promising direction for future work.

\clearpage

\begin{algorithm}[H]
\caption{Scale-Agnostic Latent Denoising (Frame-wise Bidirectional)}
\label{alg:stage1}
\begin{algorithmic}[1]
\Require LR video sequence $\mathbf{x} = \{x^i\}_{i=1}^N$, Diffusion steps $T$
\State Compute forward flows $\{f_{\text{fwd}}\}$ and backward flows $\{f_{\text{bwd}}\}$ on $\mathbf{x}$ using pre-trained RAFT
\State Initialize latents $\mathbf{z}_T = \{z_T^i\}_{i=1}^N \sim \mathcal{N}(0, \mathbf{I})$
\State $\text{reversed} \gets \text{False}$

\For{$t = T$ \textbf{to} $1$}
    \State $f \gets \{f_{\text{bwd}}\}$ \textbf{if} reversed \textbf{else} $\{f_{\text{fwd}}\}$
    \State $H_{\text{past}} \gets \emptyset$ \Comment{Initialize recurrent features}
    \For{$i = 1$ \textbf{to} $N$}
        \If{$i == 1$}
            \State $H_t^i \gets \emptyset$ \Comment{Bypass ControlNet for the first frame}
        \Else
            \State $\hat{y}_{\text{base}}^{i-1 \rightarrow i} \gets \mathrm{Warp}(\mathcal{D}(\hat{z}_0^{i-1}), f^{i-1 \rightarrow i})$ \Comment{Warp decoded RGB anchor}
            \State $H^{i-1 \rightarrow i} \gets \mathrm{Warp}(H_{\text{past}}, f^{i-1 \rightarrow i})$ \Comment{Warp past ControlNet residuals}
            \State $H_t^i \gets \mathcal{C}_\phi([z_t^i, x^i], \hat{y}_{\text{base}}^{i-1 \rightarrow i}, H^{i-1 \rightarrow i})$ \Comment{Extract gated features via TGFR}
            \State $H_{\text{past}} \gets H_t^i$ \Comment{Update recurrent features for the next frame}
        \EndIf
        \State $\epsilon_t^i \gets \epsilon_\theta([z_t^i, x^i], t, H_t^i)$ \Comment{Predict noise with frozen U-Net}
        \State $z_{t-1}^i, \hat{z}_0^i \gets \mathrm{DDPM\_Update}(z_t^i, \epsilon_t^i, t)$
    \EndFor
    \State Reverse the order of sequence $\mathbf{z}_{t-1}$, $\hat{\mathbf{z}}_0$, and $\mathbf{x}$ \Comment{Alternating propagation direction}
    \State $\text{reversed} \gets \textbf{not } \text{reversed}$
\EndFor
\State Restore original temporal order of $\mathbf{z}_0$ and $\mathbf{x}$
\State \Return Clean latent sequence $\mathbf{z}_0 = \{z_0^i\}_{i=1}^N$, Flows $\{f_{\text{fwd}}\}$, $\{f_{\text{bwd}}\}$
\end{algorithmic}
\end{algorithm}

\begin{algorithm}[H]
\caption{Continuous Arbitrary-Scale Decoding (Sequence-level)}
\label{alg:stage2}
\begin{algorithmic}[1]
\Require Clean latent sequence $\mathbf{z}_0 = \{z_0^i\}_{i=1}^N$, LR sequence $\mathbf{x}$, Target scale $s$, Flows $\{f_{\text{fwd}}\}$, $\{f_{\text{bwd}}\}$
\State $s_c \gets \min(s, 4.0)$ \Comment{OOD bounding strategy for scale-aware modules}
\State Extract intermediate features $\mathbf{F} = \{F^i\}_{i=1}^N$ by passing $\mathbf{z}_0$ through VAE decoder $\mathcal{D}$
\State $\mathbf{u} \gets \text{ScaleAwareAdapter}(\mathbf{F}, s_c)$ \Comment{Spatial scale conditioning}
\State $\mathbf{u}_{\text{fwd}} \gets \text{ForwardPropagation}(\mathbf{u}, \{f_{\text{fwd}}\})$ \Comment{Forward cascaded TGFR alignment}
\State $\tilde{\mathbf{u}} \gets \text{BackwardPropagation}(\mathbf{u}_{\text{fwd}}, \{f_{\text{bwd}}\})$ \Comment{Backward cascaded TGFR alignment}
\State $\mathbf{u}_{\text{ref}} \gets \mathrm{SAFR}(\tilde{\mathbf{u}}, s_c)$ \Comment{Scale-conditioned spectral gating}
\State Generate continuous spatial coordinates $\mathbf{p}$ for target resolution $(sH) \times (sW)$
\For{$i = 1$ \textbf{to} $N$}
    \State $\Delta y^i \gets \mathrm{LIIF}(\mathbf{u}_{\text{ref}}^i, \mathbf{p})$ \Comment{Query local features using coordinates}
    \State $\hat{y}^i \gets \Delta y^i + \mathrm{Sample}_{\text{bilinear}}(x^i, \mathbf{p})$ \Comment{Continuous coordinate-based rendering}
\EndFor
\State \Return Final HR video sequence $\hat{\mathbf{y}} = \{\hat{y}^i\}_{i=1}^N$
\end{algorithmic}
\end{algorithm}

\begin{table*}[p]
\centering
\caption{Comprehensive quantitative comparison with state-of-the-art methods across diverse scales ($2\times$, $2.5\times$, $3.25\times$, $3.5\times$, $6\times$, $8\times$) on the REDS4 dataset. For better readability, LPIPS, DISTS, and tOF are scaled by $10^2$, and tLPIPS is scaled by $10^3$. The best and second-best results are highlighted in \textbf{\color{red}{red}} and \underline{\color{blue}{blue}}, respectively.}
\label{tab:supp_master_table}
\vspace{-2mm}
\resizebox{\linewidth}{!}{
\begin{tabular}{l | cccccc | cccccc}
\toprule
\multirow{2}{*}{\textbf{Method}} & \multicolumn{6}{c|}{\textbf{$2\times$}} & \multicolumn{6}{c}{\textbf{$2.5\times$}} \\
\cmidrule(lr){2-7} \cmidrule(lr){8-13}
& LPIPS$\downarrow$ & DISTS$\downarrow$ & PSNR$\uparrow$ & SSIM$\uparrow$ & tLPIPS$\downarrow$ & tOF$\downarrow$ & LPIPS$\downarrow$ & DISTS$\downarrow$ & PSNR$\uparrow$ & SSIM$\uparrow$ & tLPIPS$\downarrow$ & tOF$\downarrow$ \\
\midrule
\multicolumn{13}{c}{Arbitrary-scale Regression-based VSR} \\
\midrule
VideoINR~\cite{chen2022videoinr}      & 12.26 & 5.49 & 24.87 & 0.7346 & 9.22 & 64.41 & 14.42 & 6.67 & 26.42 & 0.7940 & 7.21 & 52.91 \\
MoTIF~\cite{chen2023motif}         & 8.39 & 4.08 & 32.36 & 0.9269 & 9.23 & 42.46 & 12.43 & 5.29 & 31.85 & 0.9110 & 8.11 & 23.61 \\
ST-AVSR~\cite{shang2024stavsr}       & 17.42 & 6.43 & 32.51 & 0.9259 & 15.61 & 14.04 & 24.02 & 9.33 & 30.53 & 0.8851 & 18.68 & 16.61 \\
SAVSR~\cite{li2024savsr}         & 6.66 & 2.95 & 35.25 & 0.9523 & \underline{\color{blue}{7.50}} & \underline{\color{blue}{7.82}} & 11.34 & 4.90 & \textbf{\color{red}{33.01}} & \textbf{\color{red}{0.9257}} & \underline{\color{blue}{6.85}} & \underline{\color{blue}{11.33}} \\
BF-STVSR~\cite{kim2025bfstvsr}      & 6.43 & 3.15 & 34.74 & 0.9503 & 9.14 & 11.17 & \underline{\color{blue}{10.86}} & 4.90 & \underline{\color{blue}{32.75}} & \underline{\color{blue}{0.9225}} & 6.91 & 14.02 \\
V$^3$VSR~\cite{becker2025v3vsr}      & 6.01 & 2.67 & \textbf{\color{red}{36.13}} & \textbf{\color{red}{0.9592}} & 7.75 & 7.95 & 12.12 & \underline{\color{blue}{4.78}} & 32.37 & 0.9209 & 7.09 & 12.82 \\
\midrule
\multicolumn{13}{c}{Fixed-scale Generative VSR + Bicubic} \\
\midrule
RealBasicVSR~\cite{chan2022realbasicvsr}  & 8.26 & 4.88 & 31.16 & 0.9051 & 9.12 & 22.98 & 19.89 & 5.46 & 28.92 & 0.8547 & 8.45 & 25.22 \\
Upscale-A-Video~\cite{zhou2024upscale}& 31.27 & 15.03 & 26.32 & 0.7428 & 17.75 & 57.35 & 35.37 & 17.11 & 25.42 & 0.7049 & 19.74 & 68.99 \\
MGLD-VSR~\cite{yang2024mgldvsr}      & 8.70 & 4.90 & 30.86 & 0.8939 & 9.39 & 22.96 & 15.36 & 5.45 & 28.61 & 0.8358 & 9.92 & 29.54 \\
StableVSR~\cite{rota2024stablevsr}     & \underline{\color{blue}{5.51}} & \underline{\color{blue}{1.99}} & 33.49 & 0.9328 & 8.25 & 8.01 & 14.95 & 5.15 & 30.59 & 0.8792 & 8.54 & 13.09 \\
STAR~\cite{xie2025star}          & 26.37 & 12.21 & 25.08 & 0.7231 & 43.59 & 45.52 & 22.26 & 10.53 & 23.97 & 0.6892 & 23.70 & 56.40 \\
\midrule
\multicolumn{13}{c}{Arbitrary-scale Generative VSR} \\
\midrule
VEnhancer~\cite{he2024venhancer}    & 23.39 & 10.00 & 23.27 & 0.6731 & 22.28 & 77.78 & 26.88 & 11.36 & 23.41 & 0.6749 & 25.89 & 78.86 \\
\midrule
\textbf{Ours} & \textbf{\color{red}{3.84}} & \textbf{\color{red}{1.72}} & \underline{\color{blue}{35.47}} & \underline{\color{blue}{0.9534}} & \textbf{\color{red}{3.23}} & \textbf{\color{red}{7.37}} & \textbf{\color{red}{4.13}} & \textbf{\color{red}{2.04}} & 32.60 & 0.9158 & \textbf{\color{red}{3.67}} & \textbf{\color{red}{9.85}} \\
\bottomrule
\end{tabular}
}

\vspace{2mm}

\resizebox{\linewidth}{!}{
\begin{tabular}{l | cccccc | cccccc}
\toprule
\multirow{2}{*}{\textbf{Method}} & \multicolumn{6}{c|}{\textbf{$3.25\times$}} & \multicolumn{6}{c}{\textbf{$3.5\times$}} \\
\cmidrule(lr){2-7} \cmidrule(lr){8-13}
& LPIPS$\downarrow$ & DISTS$\downarrow$ & PSNR$\uparrow$ & SSIM$\uparrow$ & tLPIPS$\downarrow$ & tOF$\downarrow$ & LPIPS$\downarrow$ & DISTS$\downarrow$ & PSNR$\uparrow$ & SSIM$\uparrow$ & tLPIPS$\downarrow$ & tOF$\downarrow$ \\
\midrule
\multicolumn{13}{c}{Arbitrary-scale Regression-based VSR} \\
\midrule
VideoINR~\cite{chen2022videoinr}      & 21.96 & 9.11 & 24.24 & 0.7244 & 7.04 & 35.42 & 24.50 & 9.99 & 23.17 & 0.6907 & 7.50 & 32.42 \\
MoTIF~\cite{chen2023motif}         & 21.38 & 8.20 & 23.02 & 0.6801 & \underline{\color{blue}{6.08}} & 22.03 & 21.44 & 8.78 & \textbf{\color{red}{30.00}} & \textbf{\color{red}{0.8619}} & \underline{\color{blue}{6.26}} & \underline{\color{blue}{18.53}} \\
ST-AVSR~\cite{shang2024stavsr}       & 32.55 & 13.26 & \underline{\color{blue}{27.03}} & \underline{\color{blue}{0.7888}} & 20.59 & 21.57 & 34.77 & 14.35 & 26.93 & 0.7808 & 21.50 & 21.10 \\
SAVSR~\cite{li2024savsr}         & 19.51 & 7.78 & \textbf{\color{red}{27.13}} & \textbf{\color{red}{0.8135}} & 7.57 & \underline{\color{blue}{18.68}} & 21.41 & 8.63 & \underline{\color{blue}{27.40}} & \underline{\color{blue}{0.8181}} & 8.11 & 19.13 \\
BF-STVSR~\cite{kim2025bfstvsr}      & 19.85 & 7.89 & 25.55 & 0.7659 & 6.37 & 20.13 & 22.26 & 8.85 & 24.22 & 0.7185 & 6.46 & 20.31 \\
V$^3$VSR~\cite{becker2025v3vsr}      & 18.17 & 7.42 & 26.23 & 0.7872 & 9.71 & 20.01 & 19.81 & 8.17 & 27.18 & 0.8070 & 9.86 & 21.83 \\
\midrule
\multicolumn{13}{c}{Fixed-scale Generative VSR + Bicubic} \\
\midrule
RealBasicVSR~\cite{chan2022realbasicvsr}  & \underline{\color{blue}{13.03}} & \underline{\color{blue}{5.78}} & 24.47 & 0.7178 & 6.92 & 30.34 & 18.68 & 8.11 & 25.13 & 0.7407 & 7.36 & 31.39 \\
Upscale-A-Video~\cite{zhou2024upscale}& 40.78 & 19.94 & 23.47 & 0.6376 & 23.08 & 89.66 & 42.15 & 20.75 & 23.56 & 0.6365 & 24.76 & 96.21 \\
MGLD-VSR~\cite{yang2024mgldvsr}      & 13.94 & 6.15 & 24.30 & 0.6969 & 14.52 & 32.49 & \underline{\color{blue}{14.71}} & \underline{\color{blue}{6.68}} & 24.88 & 0.7163 & 14.98 & 33.31 \\
StableVSR~\cite{rota2024stablevsr}     & 14.38 & 6.00 & 24.91 & 0.7127 & 7.22 & 20.85 & 14.94 & 7.30 & 24.98 & 0.7083 & 9.89 & 26.71 \\
STAR~\cite{xie2025star}          & 25.44 & 10.87 & 21.18 & 0.5793 & 22.00 & 67.16 & 27.37 & 11.23 & 21.24 & 0.5867 & 27.96 & 64.82 \\
\midrule
\multicolumn{13}{c}{Arbitrary-scale Generative VSR} \\
\midrule
VEnhancer~\cite{he2024venhancer}    & 30.71 & 12.95 & 22.53 & 0.6314 & 26.40 & 83.04 & 32.82 & 14.91 & 22.60 & 0.6322 & 28.78 & 89.74 \\
\midrule
\textbf{Ours} & \textbf{\color{red}{8.17}} & \textbf{\color{red}{3.21}} & 26.50 & 0.7833 & \textbf{\color{red}{3.56}} & \textbf{\color{red}{15.23}} & \textbf{\color{red}{8.83}} & \textbf{\color{red}{3.60}} & 26.71 & 0.7894 & \textbf{\color{red}{3.79}} & \textbf{\color{red}{15.65}} \\
\bottomrule
\end{tabular}
}

\vspace{2mm}

\resizebox{\linewidth}{!}{
\begin{tabular}{l | cccccc | cccccc}
\toprule
\multirow{2}{*}{\textbf{Method}} & \multicolumn{6}{c|}{\textbf{$6\times$}} & \multicolumn{6}{c}{\textbf{$8\times$}} \\
\cmidrule(lr){2-7} \cmidrule(lr){8-13}
& LPIPS$\downarrow$ & DISTS$\downarrow$ & PSNR$\uparrow$ & SSIM$\uparrow$ & tLPIPS$\downarrow$ & tOF$\downarrow$ & LPIPS$\downarrow$ & DISTS$\downarrow$ & PSNR$\uparrow$ & SSIM$\uparrow$ & tLPIPS$\downarrow$ & tOF$\downarrow$ \\
\midrule
\multicolumn{13}{c}{Arbitrary-scale Regression-based VSR} \\
\midrule
VideoINR~\cite{chen2022videoinr}      & 37.35 & 17.07 & 22.13 & 0.6266 & 7.91 & 69.21 & 45.31 & 21.31 & 23.31 & 0.6350 & 13.09 & 110.73 \\
MoTIF~\cite{chen2023motif}         & 35.65 & 16.56 & \textbf{\color{red}{26.79}} & \textbf{\color{red}{0.7490}} & 6.99 & \underline{\color{blue}{29.75}} & 43.63 & 20.72 & 25.36 & 0.6898 & 8.79 & 43.26 \\
ST-AVSR~\cite{shang2024stavsr}       & 52.48 & 22.60 & 23.59 & 0.6442 & 39.44 & 34.18 & 59.82 & 27.28 & 24.00 & 0.6311 & 51.80 & 43.48 \\
SAVSR~\cite{li2024savsr}    & 36.13 & 15.79 & \underline{\color{blue}{23.60}} & \underline{\color{blue}{0.6699}} & 10.24 & 33.35 & 43.02 & 19.53 & \underline{\color{blue}{25.50}} & \underline{\color{blue}{0.6908}} & 14.97 & 46.12 \\
BF-STVSR~\cite{kim2025bfstvsr} & 35.39 & 16.49 & 22.59 & 0.6444 & \underline{\color{blue}{6.83}} & 32.44 & 41.02 & 20.19 & 25.38 & 0.6893 & \underline{\color{blue}{7.59}} & \underline{\color{blue}{41.85}} \\
V$^3$VSR~\cite{becker2025v3vsr} & 35.79 & 15.52 & 23.42 & 0.6738 & 14.61 & 31.10 & 44.25 & 19.69 & \textbf{\color{red}{25.89}} & \textbf{\color{red}{0.7034}} & 18.25 & 44.51 \\
\midrule
\multicolumn{13}{c}{Fixed-scale Generative VSR + Bicubic} \\
\midrule
RealBasicVSR~\cite{chan2022realbasicvsr}  & 32.89 & 17.97 & 22.47 & 0.6236 & 9.33 & 49.12 & 36.00 & 16.18 & 24.14 & 0.6440 & 9.83 & 71.04 \\
Upscale-A-Video~\cite{zhou2024upscale}& 36.55 & 16.43 & 21.24 & 0.5576 & 15.44 & 143.10 & 45.46 & 20.11 & 21.70 & 0.5539 & 16.37 & 187.20 \\
MGLD-VSR~\cite{yang2024mgldvsr}      & 29.51 & 13.49 & 22.33 & 0.6035 & 20.99 & 127.95 & 37.61 & 15.86 & 23.49 & 0.6201 & 24.55 & 210.24 \\
StableVSR~\cite{rota2024stablevsr}     & \underline{\color{blue}{28.25}} & 16.54 & 22.53 & 0.6131 & 7.96 & 35.82 & \underline{\color{blue}{34.96}} & \underline{\color{blue}{15.59}} & 23.72 & 0.6227 & 12.34 & 44.09 \\
STAR~\cite{xie2025star}          & 42.11 & \underline{\color{blue}{15.06}} & 18.75 & 0.4657 & 25.38 & 109.99 & 52.26 & 19.16 & 18.42 & 0.4539 & 26.12 & 153.58 \\
\midrule
\multicolumn{13}{c}{Arbitrary-scale Generative VSR} \\
\midrule
VEnhancer~\cite{he2024venhancer}    & 39.89 & 17.15 & 20.29 & 0.5448 & 22.89 & 124.21 & 43.87 & 19.67 & 21.73 & 0.5721 & 27.88 & 129.42 \\
\midrule
\textbf{Ours} & \textbf{\color{red}{20.24}} & \textbf{\color{red}{9.93}} & 23.02 & 0.6434 & \textbf{\color{red}{5.90}} & \textbf{\color{red}{28.12}} & \textbf{\color{red}{29.43}} & \textbf{\color{red}{14.09}} & 24.95 & 0.6653 & \textbf{\color{red}{6.39}} & \textbf{\color{red}{39.13}} \\
\bottomrule
\end{tabular}
}
\end{table*}

\end{document}